%% file: main.tex
\documentclass{article}

\PassOptionsToPackage{numbers, compress}{natbib}

\usepackage[final]{neurips_2022}

\usepackage[utf8]{inputenc} 
\usepackage[T1]{fontenc}  
\usepackage{cancel}
\usepackage{booktabs}

\usepackage{hyperref}       
\usepackage{url}            
\usepackage{booktabs}       
\usepackage{amsfonts,amsmath,subcaption}       
\usepackage{enumitem,bbm, graphicx}
  \setlist{leftmargin=*,noitemsep}
 \usepackage{nicefrac}       
\usepackage{microtype}  

\usepackage{algorithm,algpseudocode}
\usepackage{cancel}
\usepackage{YK}

\def\hSigma{\widehat{\Sigma}}
\def\htheta{\widehat{\theta}}
\def\hY{\widehat{Y}}

\def\hP{\widehat{P}}

\title{Calibrated Data-Dependent Constraints with Exact Satisfaction Guarantees}

\author{
Songkai Xue \\
Department of Statistics \\
University of Michigan \\
\texttt{sxue@umich.edu} \\
\And
Yuekai Sun \\
Department of Statistics \\
University of Michigan \\
\texttt{yuekai@umich.edu} \\
\And
Mikhail Yurochkin \\
IBM Research \\
MIT-IBM Watson AI Lab \\
\texttt{mikhail.yurochkin@ibm.com}
}

\begin{document}

\maketitle

\begin{abstract}
We consider the task of training machine learning models with data-dependent constraints. Such constraints often arise as empirical versions of expected value constraints that enforce fairness or stability goals. We reformulate data-dependent constraints so that they are \emph{calibrated}: enforcing the reformulated constraints guarantees that their expected value counterparts are satisfied with a user-prescribed probability. The resulting optimization problem is amendable to standard stochastic optimization algorithms, and we demonstrate the efficacy of our method on a fairness-sensitive classification task where we wish to guarantee the classifier's fairness (at test time).
\end{abstract}

\input{sections/intro.tex}
\input{sections/methods.tex}
\input{sections/simulation}

\input{sections/fairness.tex}
\input{sections/summary.tex}

\begin{ack}
This paper is based upon work supported by the National Science Foundation (NSF) under grants no.\ 1916271, 2027737, and 2113373. 
\end{ack}

\bibliographystyle{plainnat}
\bibliography{YK, SK}

\newpage
\section*{Checklist}


\begin{enumerate}

\item For all authors...
\begin{enumerate}
  \item Do the main claims made in the abstract and introduction accurately reflect the paper's contributions and scope?
    \answerYes{}
  \item Did you describe the limitations of your work?
    \answerYes{}
  \item Did you discuss any potential negative societal impacts of your work?
    \answerNA{}
  \item Have you read the ethics review guidelines and ensured that your paper conforms to them?
    \answerYes{}
\end{enumerate}

\item If you are including theoretical results...
\begin{enumerate}
  \item Did you state the full set of assumptions of all theoretical results?
    \answerYes{}
        \item Did you include complete proofs of all theoretical results?
    \answerYes{}
\end{enumerate}

\item If you ran experiments...
\begin{enumerate}
  \item Did you include the code, data, and instructions needed to reproduce the main experimental results (either in the supplemental material or as a URL)?
    \answerYes{}
  \item Did you specify all the training details (e.g., data splits, hyperparameters, how they were chosen)?
    \answerYes{}
        \item Did you report error bars (e.g., with respect to the random seed after running experiments multiple times)?
    \answerYes{}
        \item Did you include the total amount of compute and the type of resources used (e.g., type of GPUs, internal cluster, or cloud provider)?
    \answerNo{}
\end{enumerate}

\item If you are using existing assets (e.g., code, data, models) or curating/releasing new assets...
\begin{enumerate}
  \item If your work uses existing assets, did you cite the creators?
    \answerYes{}
  \item Did you mention the license of the assets?
    \answerNA{}
  \item Did you include any new assets either in the supplemental material or as a URL?
    \answerNA{}
  \item Did you discuss whether and how consent was obtained from people whose data you're using/curating?
    \answerNA{}
  \item Did you discuss whether the data you are using/curating contains personally identifiable information or offensive content?
    \answerNA{}
\end{enumerate}

\item If you used crowdsourcing or conducted research with human subjects...
\begin{enumerate}
  \item Did you include the full text of instructions given to participants and screenshots, if applicable?
    \answerNA{}
  \item Did you describe any potential participant risks, with links to Institutional Review Board (IRB) approvals, if applicable?
    \answerNA{}
  \item Did you include the estimated hourly wage paid to participants and the total amount spent on participant compensation?
    \answerNA{}
\end{enumerate}

\end{enumerate}



\clearpage
\appendix
\title{Supplementary Materials for \\Calibrated Data-Dependent Constraints with Exact
Satisfaction Guarantees}
\author{
Songkai Xue \\
Department of Statistics \\
University of Michigan \\
\texttt{sxue@umich.edu} \\
\And
Yuekai Sun \\
Department of Statistics \\
University of Michigan \\
\texttt{yuekai@umich.edu} \\
\And
Mikhail Yurochkin \\
IBM Research \\
MIT-IBM Watson AI Lab \\
\texttt{mikhail.yurochkin@ibm.com}
}
\makeatletter
\@maketitle
\makeatother

\begin{abstract}
In Section \ref{sec:proofs-for-section-2}, \ref{sec:proof-thm-3.1}, \ref{sec:proof-cor-3.2}, we present proofs of theoretical results. In Section \ref{sec:dual-ascent-multiple-constraints}, we summarize the dual ascent algorithm for solving \eqref{eq:robust-SAA-multiple-expected-value-constraint}. In Section \ref{sec:details-simulations}, we provide details for simulations on the multi-item newsvendor problem with independent constraints and run additional simulations. In Section \ref{sec:instantiate-EO}, we demonstrate how our method can be applied to $\varepsilon$-equality of opportunity. In Section \ref{sec:details-unknown-active-set}, we show the theoretical properties of the two-stage method for unknown active set. In Section \ref{sec:proxy-dual-ascent-algorithm}, we summarize the proxy dual ascent algorithm for handling non-differentiable constraints. In Section \ref{sec:details-experiments}, we provide details for Adult experiments.
In Section \ref{sec:standard-dro-dual}, we provide a standard derivation for the dual form of the robust constraint function \eqref{eq:robust-expected-value-constraint-dual-form}.
In Section \ref{sec:more-baseline}, experiments on additional baseline and dataset are conducted.
\end{abstract}

\input{sections/appendix.tex}

\end{document}

%% file: sections/intro.tex
\section{Motivation} 
\label{sec:intro}

In machine learning (ML) practice, accuracy is often only one of many training objectives. For example, algorithmic fairness considerations may require a credit scoring system to perform comparably on men and women.
Here are a few other examples.

\paragraph{Churn rate and stability} The churn rate of an ML model compared to another model is the fraction of samples on which the predictions of the two models differ \cite{goh2016Satisfying,milanifard2016Launch}. In ML practice, one may wish to control the churn rate between a new model and its predecessor because a high churn rate can disorient users and downstream system components. One way of training models with small churn is to enforce a churn rate constraint during training.

\paragraph{Precision, recall, \etc} Classification and information retrieval models must often balance precision and recall. To train such models, practitioners carefully trade off one metric for the other by optimizing for one metric subject to constraints on the other. 

\paragraph{Resource constraints} Practitioners sometimes wish to control how often a classifier predicts a certain class due to budget or resource constraints. For example, a company that uses ML to select customers for a targeted offer may wish to constrain the fraction of customers selected for the offer. 
Another prominent example of a stochastic optimization problem with resource constraints is the newsvendor problem, which we come back to in section \ref{sec:simulation}.

Unlike constraints on the structure of model parameters (\eg, sparsity), the constraints encoding the preceding training objectives are \emph{data-dependent}. This leads to the issue of \emph{constraint generalization}: whether the constraints \emph{generalize} out-of-sample. For example, if a classifier is trained to have comparable accuracy on two subpopulations in the training data, will it also have comparable accuracy on samples from the two subpopulations at test time?

In this paper, we consider the out-of-sample generalization of \emph{expected-value} constraints. To keep things simple, consider a stochastic optimization problem with a single \emph{expected-value} constraint:
\begin{equation}
\theta^\star\in\left\{\begin{aligned}
&\argmin_{\theta\in\Theta} &&\bbE_{P_0}\big[f(\theta;Z)\big] = \textstyle\int_\cZ f(\theta;z)dP_0(z) \\
&\subjectto             &&\bbE_{P_0}\big[g(\theta;Z)\big] = \textstyle\int_\cZ g(\theta;z)dP_0(z) \leq 0
\end{aligned}\right\},
\label{eq:single-expected-value-constraint}
\end{equation}
where $\Theta$ is a (finite-dimensional) parameter space,  $f, g:\Theta\times\cZ\to\bbR$ are (known) cost and constraint functions, and $Z\in\cZ$ is a random variable that represents a sample. The distribution of $Z$ is unknown, so we cannot solve \eqref{eq:single-expected-value-constraint} directly. Instead, we obtain IID training samples $\{Z_i\}_{i=1}^n$ from the true underlying distribution $P_0$ and solve the empirical version of \eqref{eq:single-expected-value-constraint}:
\begin{equation}
\htheta_n\in\left\{\begin{aligned}
&\argmin_{\theta\in\Theta} &&\textstyle\frac1n\sum_{i=1}^n f(\theta;Z_i) \\
&\subjectto             &&\textstyle\frac1n\sum_{i=1}^n g(\theta;Z_i) \leq 0
\end{aligned}\right\}.
\label{eq:SAA-single-expected-value-constraint}
\end{equation}
The estimator $\htheta_n$ (of $\theta^\star$) is guaranteed to satisfy the empirical constraint (\ie, $\frac1n\sum_{i=1}^n g(\htheta_n;Z_i) \leq 0$), but it is unclear whether $\htheta_n$ satisfies the actual (population) constraint $\bbE_{P_0}\big[g(\theta;Z)\big] \leq 0$. As we shall see, under standard assumptions on \eqref{eq:single-expected-value-constraint}, $\htheta_n$ only satisfies the actual constraint with probability approaching $\frac12$ (see corollary \ref{cor:robust-SAA-single-expected-value-constraint-asymptotically-exact-constraint-satisfaction}). This is especially problematic for constraints that encode algorithmic fairness goals. For example, the 80\% rule published by the US Equal Employment Opportunity Commission, interpreted in the machine learning context, requires the rate at which a classifier predicts the advantaged label in minority groups to be at least 80\% of the rate at which the classifier predicts the advantaged label in the majority group \cite{barocas2019Fairness}.

In this paper, we propose a distributionally robust version of \eqref{eq:SAA-single-expected-value-constraint} that \emph{guarantees} the actual constraint $\bbE_{P_0}\big[g(\theta;Z)\big] \leq 0$ will be satisfied with probability $1-\alpha$:
\begin{equation}
\htheta_n\in\left\{\begin{aligned}
&\argmin_{\theta\in\Theta} &&\textstyle\frac1n\sum_{i=1}^nf(\theta;Z_i) \\
&\subjectto             &&\sup\nolimits_{P:D_\varphi(P\|\hP_n)\leq\frac{\rho_\alpha}{n}}\bbE_{P}\big[g(\theta;Z)\big] \leq 0
\end{aligned}\right\},
\label{eq:robust-SAA-single-expected-value-constraint}
\end{equation}
where $D_\varphi$ is a $\varphi$-divergence (see section \ref{sec:single-expected-value-constraint} for details), $\hP_n$ is the empirical distribution of the training samples, and $\sqrt{\rho_\alpha}$ is the $1-\alpha$ quantile of a standard normal random variable. More concretely, we show that $\htheta_n$ achieves \emph{asymptotically exact constraint satisfaction}
\begin{equation}
\lim_{n\to\infty}\bbP\left\{\bbE_{P_0}\big[g(\htheta_n;Z)\big] \leq 0\right\} = 1-\alpha.
\label{eq:asymptotically-exact-constraint-satisfaction}
\end{equation}
Here the inner expectation is with respect to $Z$; the outer probability is with respect to the training samples $\{Z_i\}_{i=1}^n$. Three desirable properties of \eqref{eq:robust-SAA-single-expected-value-constraint} are
\begin{enumerate}
\item \textbf{exact constraint satisfaction:} If the actual probability of constraint satisfaction exceeds $1-\alpha$, then the method is too conservative. This may (unnecessarily) increase the cost of the model. By picking $\rho_\alpha$ in \eqref{eq:robust-SAA-single-expected-value-constraint} carefully, constraints are satisfied with asymptotically exact probability $1-\alpha$.
\item \textbf{computationally efficient:} As we shall see, the computational cost of solving \eqref{eq:robust-SAA-single-expected-value-constraint} is comparable to the cost of solving distributionally robust sample average approximation (SAA) problems. 
\item \textbf{pivotal:} There are no nuisance parameters to estimate (\eg, asymptotic variances) in \eqref{eq:robust-SAA-single-expected-value-constraint}. The user merely needs to look up the correct quantile of the standard normal distribution for their desired level of constraint generalization.
\end{enumerate}
The rest of this paper is organized as follows.
In Section \ref{sec:single-expected-value-constraint}, we develop method, theory, and algorithm for stochastic optimization problems with single constraint. In Section \ref{sec:multiple-expected-value-constraints}, we extend our method, theory, and algorithm to stochastic optimization problems with multiple constraints.
In Section \ref{sec:simulation}, we validate our theory by simulating a resource-constrained newsvendor problem.
In Section \ref{sec:fairness}, we demonstrate the efficacy of our method by using it to train an algorithmically fair income classifier.
In addition, we show how to apply our method to a fairness constrained learning problem and discuss two practical considerations for fair ML application scenarios.
Finally, we summarize our work in Section \ref{sec:summary} and point out an interesting avenue of future work.

\subsection{Related work}

The closest work to our work is \cite{lam2019Recovering}.
They seek to pick a (data-dependent) \emph{uncertainty set} $\cU$ such that
\begin{equation}
\lim_{n\to\infty}\bbP\left\{\sup\nolimits_\theta\left\{\bbE_{P_0}\big[g(\theta;Z)\big] - \sup\nolimits_{P\in\cU}\bbE_P\big[g(\theta;Z)\big]\right\} \leq 0\right\} = 1-\alpha.
\label{eq:uniform-converage}
\end{equation}
This condition is stronger than necessary: we only require 
\begin{equation}
\lim_{n\to\infty}\bbP\left\{\bbE_{P_0}\big[g(\htheta_n;Z)\big] - \sup\nolimits_{P\in\cU}\bbE_P\big[g(\htheta_n;Z)\big] \leq 0\right\} =1-\alpha
\end{equation}
where $\htheta_n$ is a (data-dependent) estimator (not necessarily \eqref{eq:SAA-single-expected-value-constraint} or \eqref{eq:robust-SAA-single-expected-value-constraint}).
\cite{lam2019Recovering} study (asymptotic) constraint satisfaction \eqref{eq:asymptotically-exact-constraint-satisfaction} for all deterministic objective functions (see \cite{lam2019Recovering}, \S 1.1 for details).
They advocate picking a KL divergence ball with radius that depends on the excursion probability of a certain $\chi^2$ process. 

Another closely related line of work is on data-splitting approaches for ensuring constraint generalization \cite{woodworth2017Learning,cotter2018Training}. At a high level, they split the training data into a training and validation subsets and use the validation subset to tune models trained on the training subset so that they satisfy the constraints. Although (computationally) simple and intuitive, their approach does not allow users to precisely control the constraint violation probability. 

\cite{lam2019Recovering} is the latest in a line of work on distributionally robust optimization (DRO) that show the optimal value of DRO problems 
\begin{equation}
\min\nolimits_{\theta\in\Theta}\sup\nolimits_{P\in\cU}\bbE_P\big[g(\theta;Z)\big],
\label{eq:DRO}
\end{equation}
where $\cU$ is a (data-dependent) uncertainty set of probability distributions, are upper confidence bounds for the optimal values of stochastic optimization problems. Common choices of uncertainty sets in DRO include uncertainty sets defined by moment or support constraints \cite{chen2007Robust,delage2010Distributionally,goh2010Distributionally}, $\varphi$-divergences \cite{ben-tal2012Robust,lam2015Quantifying,namkoong2016Stochastic}, and Wasserstein distances \cite{shafieezadeh-abadeh2015Distributionally,blanchet2016Robust,esfahani2015Datadriven,lee2017Minimax,sinha2017Certifying}. This line of work is motivated by Owen's seminal work on empirical likelihood \cite{owen1990Empirical}. In recent work, \cite{lam2015Quantifying,duchi2016Statistics} show that the optimal value of DRO problems with empirical likelihood uncertainty sets leads to asymptotically exact upper confidence bounds for the optimal value of stochastic optimization problems (\cite{duchi2016Statistics} consider more general $\varphi$-divergence uncertainty sets). \cite{blanchet2016Robust} establish similar coverage results for Wasserstein uncertainty sets. 

Our work is also closely related to the work on the variance regularization properties of DRO \cite{namkoong2016Stochastic}, which uses DRO to approximate the variance regularization cost function (see \eqref{eq:variance-regularization-DRO-approximation}). \cite{gao2017Robust} establish similar results for Wasserstein DRO.
Lastly, we relate our work to the literature on chance constrained optimization (see \cite{jiang2016data} and the references therein). The general goal of chance constrained optimization is to minimize a loss function subject to the probability of satisfying uncertain constraints is above a prescribed level. While our methods reformulate expected value constraints and we show that the solution of the reformulated problem enjoys an asymptotically exact probabilistic guarantees of constraint satisfaction. In addition, the data-dependent constraints in our work are also unknown in practice, which differs from the common setup in the chance constrained optimization literature.

%% file: sections/methods.tex
\section{Single expected value constraint}
\label{sec:single-expected-value-constraint}

We motivate \eqref{eq:robust-SAA-single-expected-value-constraint} by considering a few alternatives. First, we note that the results later in this section show that \eqref{eq:SAA-single-expected-value-constraint} violates the actual constraint in \eqref{eq:single-expected-value-constraint} approximately half the time (see corollary \ref{cor:robust-SAA-single-expected-value-constraint-asymptotically-exact-constraint-satisfaction}). The most straightforward modification of \eqref{eq:SAA-single-expected-value-constraint} to ensure $\htheta_n$ satisfies the (actual) constraint $\bbE_{P_0}\big[g(\theta;Z)\big] \leq 0$ is to add a ``margin'' in \eqref{eq:robust-SAA-single-expected-value-constraint}; \ie\ enforce the constraint
\begin{equation}\textstyle
\frac1n\sum_{i=1}^n g(\theta;Z_i) + \eps_n\le 0
\label{eq:uniform-margin-constraint}
\end{equation}
in \eqref{eq:SAA-single-expected-value-constraint}. If we pick the slack term $\eps_n$ such that
\[\textstyle
\bbP\left\{\sup\nolimits_{\theta\in\Theta}\left\{\bbE_{P_0}\big[g(\theta;Z)\big] - \frac1n\sum_{i=1}^n g(\theta;Z_i)\right\} > \eps_n\right\} \leq \alpha,
\]
then it is not hard to check that the resulting $\htheta_n$ satisfies the (actual) constraint with probability greater than $1-\alpha$ \cite{wang2008Sample,luedtke2008Sample}. However, this approach is most likely conservative because the constraint is unnecessarily stringent for $\theta$'s such that $\frac1n\sum_{i=1}^ng(\theta;Z_i)$ is less variable. It is also not pivotal: $\eps_n$ is often set using bounds from (uniform) concentration inequalities, which typically depend on unknown problem parameters.

To relax the empirical constraint in a way that adapts to the variability of the empirical constraints, we replace the uniform margin in \eqref{eq:uniform-margin-constraint} with a parameter-dependent margin:
\begin{equation}\textstyle
\frac1n\sum_{i=1}^n g(\theta;Z_i) + z_\alpha\frac{\hsigma(\theta)}{\sqrt{n}}\le 0,
\label{eq:parameter-dependent-margin-constraint}
\end{equation}
where $z_\alpha$ is the $1-\alpha$ quantile of a standard normal random variable and $\hsigma^2(\theta)$ is an estimate of the asymptotic variance of $g(\theta;Z)$. We recognize the (parameter-dependent) margin as (a multiple of) the standard error of the empirical constraint. It is possible to show that enforcing \eqref{eq:parameter-dependent-margin-constraint} achieves asymptotically exact constraint generalization \eqref{eq:asymptotically-exact-constraint-satisfaction} \cite{lam2019Recovering}.

The main issue with this method is it is not amenable to standard stochastic optimization algorithms. In particular, even if the original constraint in \eqref{eq:SAA-single-expected-value-constraint} is convex, \eqref{eq:parameter-dependent-margin-constraint} is generally non-convex.
Another issue is that it is not pivotal: the user must estimate the asymptotic variance of $g(\theta;Z)$. 

To overcome these two issues, we consider a distributionally robust version of \eqref{eq:SAA-single-expected-value-constraint}; \ie\ enforcing 
\begin{equation}
\sup\nolimits_{P:D_\varphi(P\| \hP_n)\leq\frac{\rho_\alpha}{n}}\bbE_P\big[g(\theta;Z)\big] \leq 0,
\label{eq:DRO-constraint}
\end{equation}
where $D_\varphi(P\|Q) \triangleq \int\varphi(\frac{dP}{dQ})dQ$ is a $\varphi$-divergence. Common choices of $\varphi$ include $\varphi(t) = (t-1)^2$ (which leads to the  $\chi^2$-divergence) and $\varphi(t) = -\log t + t-1$ (which leads to the Kullback-Leibler divergence).
Although there are many other choices for the uncertainty set in \eqref{eq:DRO-constraint}, we pick an $\varphi$-divergence ball because (i) \eqref{eq:DRO-constraint} with an $\varphi$-divergence ball is asymptotically equivalent to \eqref{eq:parameter-dependent-margin-constraint}:
\begin{equation}
\sup\nolimits_{P:D_\varphi(P\| \hP_n)\le\frac{\rho_\alpha}{n}}\bbE_P\big[g(\theta;Z)\big] \textstyle\approx \frac1n\sum_{i=1}^n g(\theta;Z_i) + z_{\alpha}\frac{\hsigma(\theta)}{\sqrt{n}},
\label{eq:variance-regularization-DRO-approximation}
\end{equation}
and (ii) it leads to pivotal uncertainty sets. For theoretical analysis, we always use $\varphi(t) = (t-1)^2$ and $\chi^2$-divergence in the remainder of this paper.

Before we state the asymptotically exact constraint satisfaction property of \eqref{eq:robust-SAA-single-expected-value-constraint} rigorously, we describe our assumptions on the problem.
\begin{enumerate}
\item \textbf{smoothness and concentration:} $f$ and $g$ are twice continuously differentiable with respect to $\theta$, and $f(\theta^\star;Z)$, $\nabla f(\theta^\star;Z)$, $g(\theta^\star;Z)$, $\nabla g(\theta^\star;Z)$ are sub-Gaussian random variables. 
\item \textbf{uniqueness:} the stochastic optimization problem with a single expected value constraint \eqref{eq:single-expected-value-constraint} has a unique optimal primal-dual pair $(\theta^\star,\lambda^\star)$, and $\theta^\star$ belongs to the interior of the compact set $\Theta$. 
\item \textbf{strict complementarity:} $\lambda^\star > 0$.
\item \textbf{positive definiteness:} The Hessian of the Lagrangian evaluated at  $(\theta^\star,\lambda^\star)$ is positive definite.
\end{enumerate}

The preceding assumptions are not the most general, but they are easy to interpret. The smoothness conditions on $f$ and $g$ with respect to $\theta$, the concentration conditions of $f(\theta^\star;Z)$ and $g(\theta^\star;Z)$, and the uniqueness condition facilitate the use of standard tools from asymptotic statistics to study the large sample properties of the constraint value. The strict complementarity condition rules out problems in which the constraint is extraneous; \ie\ problems in which the unconstrained minimum coincides with the constrained minimum.

We are ready to state the asymptotically exact constraint satisfaction property of \eqref{eq:robust-SAA-single-expected-value-constraint} rigorously. The main technical result characterizes the limiting distribution of the constraint value.

\begin{theorem}
\label{thm:robust-SAA-single-expected-value-constraint-asymptotically-exact-constraint-satisfaction}
Let $\htheta_n$ be an optimal solution of \eqref{eq:robust-SAA-single-expected-value-constraint} converging in probability as $n\to\infty$ to $\theta^\star$. Under the standing assumptions, we have
\begin{equation*}
\sqrt{n}\left(\bbE_{P_0}\big[g(\htheta_n;Z)\big] - \cancel{\bbE_{P_0}\big[g(\theta^\star;Z)\big]}\right) \dto \cN\left(-\sqrt{\rho_\alpha\Var_{P_0}[g(\theta^\star;Z)]},\Var_{P_0}[g(\theta^\star;Z)]\right).
\end{equation*}
\end{theorem}

We translate this result on the constraint value to a result on constraint generalization.

\begin{corollary}
\label{cor:robust-SAA-single-expected-value-constraint-asymptotically-exact-constraint-satisfaction}
Let $\sqrt{\rho_\alpha}$ be the $1-\alpha$ quantile of a standard normal random variable. Under the conditions of theorem \ref{thm:robust-SAA-single-expected-value-constraint-asymptotically-exact-constraint-satisfaction}, we have
\[
\lim_{n\to\infty}\bbP\left\{\bbE_{P_0}\big[g(\htheta_n;Z)\big] \leq 0\right\} = \bbP\left\{U \leq \sqrt{\rho_\alpha}\right\} = 1-\alpha,
\]
where $U\sim\cN(0,1)$ is a standard Gaussian random variable.
\end{corollary}

From theorem \ref{thm:robust-SAA-single-expected-value-constraint-asymptotically-exact-constraint-satisfaction} and corollary \ref{cor:robust-SAA-single-expected-value-constraint-asymptotically-exact-constraint-satisfaction} (see proofs in Appendix \ref{sec:proofs-for-section-2}), we find that \begin{enumerate}
\item picking $\rho_\alpha = 0$ (\ie, equivalently solving \eqref{eq:SAA-single-expected-value-constraint}) leads to a constraint violation probability that approaches $\frac12$ in the large sample limit.
\item the relation between the mean and variance of the limiting distribution of the constraint value in Theorem \ref{thm:robust-SAA-single-expected-value-constraint-asymptotically-exact-constraint-satisfaction} allows us to pick $\rho_\alpha$ in a pivotal way (\ie\ does not depend on nuisance parameters).
\end{enumerate}

\subsection{Stochastic approximation for \eqref{eq:robust-SAA-single-expected-value-constraint}}

In the rest of this section, we derive a stochastic optimization algorithm to solve \eqref{eq:robust-SAA-single-expected-value-constraint} efficiently. As we shall see, the computational cost of this algorithm is comparable to the cost of solving a DRO problem. The key insight is that the robust constraint function has a dual form (see Appendix \ref{sec:standard-dro-dual}):
\begin{equation}
\sup\nolimits_{P:D_\varphi(P\| \hP_n)\leq\rho}\bbE_P\big[g(\theta;Z)\big] \textstyle=\inf\nolimits_{\mu\ge 0,\nu\in\bbR}\left\{\frac1n\sum_{i=1}^n \mu\varphi^*\big(\frac{g(\theta;Z_i) - \nu}{\mu}\big) + \mu\rho + \nu\right\},
\label{eq:robust-expected-value-constraint-dual-form}
\end{equation}
where $\varphi^*(s) \triangleq \sup\nolimits_t \{st - \varphi(t)\}$ is the convex conjugate of $\varphi$. As we use $\chi^2$-squared divergence and $\varphi(t) = (t-1)^2$, the corresponding $\varphi^*(s) = \frac{s^2}{4} + s$. The Lagrangian of \eqref{eq:robust-SAA-single-expected-value-constraint} is 
\[
\begin{aligned}
L(\theta,\lambda) &\textstyle\triangleq \frac1n\sum_{i=1}^n f(\theta;Z_i) + \lambda \sup\nolimits_{P:D_\varphi(P\| \hP_n)\leq\frac{\rho_\alpha}{n}}\bbE_P\big[g(\theta;Z)\big] \\
&\textstyle=\frac1n\sum_{i=1}^n f(\theta;Z_i) + \lambda\textstyle\inf\nolimits_{\mu\ge 0,\nu\in\bbR}\left\{\frac1n\sum_{i=1}^n\mu\varphi^*\big(\frac{g(\theta;Z_i) - \nu}{\mu}\big) + \mu\frac{\rho_\alpha}{n} + \nu\right\}.
\end{aligned}
\]
We see that evaluating the dual function $\inf_\theta L(\theta,\lambda)$ (at a fixed $\lambda$) entails solving a stochastic optimization problem that is suitable for stochastic approximation.
This suggests a dual ascent algorithm for solving \eqref{eq:robust-SAA-single-expected-value-constraint}:
\begin{enumerate}
\item evaluate the dual function at $\lambda_t$ by solving a stochastic optimization problem. 
\item update $\lambda_t$ with a dual ascent step.
\end{enumerate}
We summarize this algorithm in Algorithm \ref{alg:robust-SAA-single-expected-value-constraint}. The main cost of Algorithm \ref{alg:robust-SAA-single-expected-value-constraint} is incurred in the third line: evaluating the dual function. Fortunately, this step is suitable for stochastic approximation, so we can leverage recent advances in the literature to reduce the (computational) cost of this step. The total cost of this algorithm is comparable to that of distributionally robust optimization.

\begin{algorithm*}
   \caption{Dual ascent algorithm for \eqref{eq:robust-SAA-single-expected-value-constraint}}
   \label{alg:robust-SAA-single-expected-value-constraint}
\begin{algorithmic}[1]
   \State {\bfseries Input:} starting dual iterate $\lambda_0 \ge 0$
   \Repeat
   \State Evaluate dual function: \[\textstyle\quad~~(\theta_t,\mu_t,\nu_t) \gets \argmin_{\theta,\mu\ge 0,\nu}\frac1n\sum_{i=1}^n f(\theta;Z_i) + \lambda_t\left\{\textstyle\frac1n\sum_{i=1}^n\mu\varphi^*\big(\frac{g(\theta;Z_i) - \nu}{\mu}\big) + \mu\frac{\rho_\alpha}{n} + \nu\right\}\]
   \State Dual ascent update: $\lambda_{t+1} \gets \left[\lambda_t + \eta_t\left\{\textstyle\frac1n\sum_{i=1}^n\mu_t\varphi^*(\frac{g(\theta_t;Z_i) - \nu_t}{\mu_t}) + \mu_t\frac{\rho_\alpha}{n} + \nu_t\right\}\right]_+$
   \Until{converged}
\end{algorithmic}
\end{algorithm*}

\section{Multiple expected value constraints}
\label{sec:multiple-expected-value-constraints}

In this section, we extend the results from the preceding section to stochastic optimization problems with multiple data-dependent constraints. Consider a stochastic optimization problem with $K$ expected value constraints 
\begin{equation}
\theta^\star\in\left\{\begin{aligned}
&\argmin_{\theta\in\Theta} &&\bbE_{P_0}\big[f(\theta;Z)\big] \\
&\subjectto             &&\left\{\bbE_{P_0}\big[g_k(\theta;Z)\big] \leq 0\right\}_{k=1}^K
\end{aligned}\right\},
\label{eq:multiple-expected-value-constraint}
\end{equation}
Following the development in Section \ref{sec:single-expected-value-constraint}, we enforce the expected value constraints with robust versions of the sample average constraints:
\begin{equation}
\htheta_n\in\left\{\begin{aligned}
&\argmin_{\theta\in\Theta} &&\textstyle\frac1n\sum_{i=1}^nf(\theta;Z_i) \\
&\subjectto             &&\left\{\sup\nolimits_{P:D_\varphi(P\| \hP_n)\le\frac{\rho_k}{n}}\bbE_P\big[g_k(\theta;Z)\big] \le 0\right\}_{k=1}^K
\end{aligned}\right\},
\label{eq:robust-SAA-multiple-expected-value-constraint}
\end{equation}
where $\brho = (\rho_1,\dots,\rho_K)^\top$ are uncertainty set radii for the constraints. 
There are other approaches to enforcing multiple constraints that result in constraint generalization; we focus on \eqref{eq:robust-SAA-multiple-expected-value-constraint} here because it allows the user to adjust the constraint generalization probability for different constraints. 

First, we extend theorem \ref{thm:robust-SAA-single-expected-value-constraint-asymptotically-exact-constraint-satisfaction} and corollary \ref{cor:robust-SAA-single-expected-value-constraint-asymptotically-exact-constraint-satisfaction} to problems with multiple (expected value) constraints.
We assume
\begin{enumerate}
\item \textbf{smoothness and concentration:} for $k\in [K]$, $f, g_k$ are twice continuously differentiable with respect to $\theta$, and $f(\theta^\star;Z), \nabla f(\theta^\star;Z)$, $g_k(\theta^\star;Z),\nabla g_k(\theta^\star;Z)$ are sub-Gaussian random variables. 
\item \textbf{uniqueness:} the stochastic optimization problem with $K$ expected value constraints \eqref{eq:multiple-expected-value-constraint} has a unique optimal primal-dual pair $(\theta^\star,\blambda^
\star)$, and $\theta^\star$ belongs to the interior of the compact set $\Theta$.
\item \textbf{strict complementarity:} $\blambda^\star\in\intr(\bbR_+^K)$, \ie, each component of $\blambda^\star$ is strictly positive.
\item \textbf{positive definiteness:} The Hessian of the Lagrangian evaluated at  $(\theta^\star,\blambda^\star)$ is positive definite.
\end{enumerate}

The strict complementarity constraint seems especially strong here because it requires all the constraints to be active. It is possible (with extra notational overhead) to state the result in terms of just the active constraints.
We refer to Section \ref{sec:unknown-active-set} for more information about the unknown active set.
Further, as long as the sample size is large enough, the active constraints in  \eqref{eq:robust-SAA-multiple-expected-value-constraint} coincide with the active constraints in \eqref{eq:multiple-expected-value-constraint}. To keep things simple, we assume all the constraints are active.

\begin{theorem}
\label{thm:robust-SAA-multiple-expected-value-constraint-asymptotically-exact-constraint-satisfaction}
Let $\htheta_n$ be an optimal solution of \eqref{eq:robust-SAA-multiple-expected-value-constraint} converging in probability as $n\to\infty$ to $\theta^\star$. Under the standing assumptions, we have
\[
\sqrt{n}\begin{bmatrix}\bbE_{P_0}\big[g_1(\htheta_n;Z)\big] \\ \vdots \\ \bbE_{P_0}\big[g_K(\htheta_n;Z)\big]\end{bmatrix} \dto \cN\left(-\begin{bmatrix}\sqrt{\rho_1\Var_{P_0}[g_1(\theta^\star;Z)]} \\ \vdots \\ \sqrt{\rho_K\Var_{P_0}[g_K(\theta^\star;Z)]}\end{bmatrix},\Var_{P_0}\begin{bmatrix}g_1(\theta^\star;Z)\\ \vdots \\ g_K(\theta^\star;Z)\end{bmatrix}\right).
\]
\end{theorem}

\begin{corollary}
\label{cor:robust-SAA-multiple-expected-value-constraint-asymptotically-exact-constraint-satisfaction}
Under the conditions of theorem \ref{thm:robust-SAA-multiple-expected-value-constraint-asymptotically-exact-constraint-satisfaction}, we have
\[
\lim_{n\to\infty}\bbP\left\{\begin{bmatrix}\bbE_{P_0}\big[g_1(\htheta_n;Z)\big] \\ \vdots \\ \bbE_{P_0}\big[g_K(\htheta_n;Z)\big]\end{bmatrix} \in -\bbR_+^K\right\} = \bbP\{\bU \le \sqrt{\brho}\},
\]
where $\sqrt{\brho} = (\sqrt{\rho_1}, \ldots, \sqrt{\rho_K})^\top$, and $\bU$ is a Gaussian random vector with mean zero and covariance 
\begin{equation}
\begin{aligned}
\Corr_{P_0}\begin{bmatrix}g_1(\theta^\star;Z)\\ \vdots \\ g_K(\theta^\star;Z)\end{bmatrix} \triangleq D^{-\frac12}\Cov_{P_0}\begin{bmatrix}g_1(\theta^\star;Z)\\ \vdots \\ g_K(\theta^\star;Z)\end{bmatrix}D^{-\frac12}, \\
D \triangleq \diag\left(\{\Var_{P_0}[g_k(\theta^\star,Z)]\}_{k=1}^K\right).
\end{aligned}
\label{eq:constraint-correlation}
\end{equation}
\end{corollary}

From theorem \ref{thm:robust-SAA-multiple-expected-value-constraint-asymptotically-exact-constraint-satisfaction} and corollary \ref{cor:robust-SAA-multiple-expected-value-constraint-asymptotically-exact-constraint-satisfaction} (see proofs in Appendix \ref{sec:proof-thm-3.1} and \ref{sec:proof-cor-3.2}), we find that the probability of constraint satisfaction decreases \emph{exponentially} as the number of constraints increases. We also see that our method is no longer pivotal for multiple expected value constraints: the uncertainty set radii depends on the (unknown) correlation structure among the constraint values. 
Fortunately, it is not hard to estimate this correlation structure. The most straightforward way is with the empirical correlation matrix. Let $\hSigma_n$ be the empirical covariance matrix of the constraint values.
The empirical correlation matrix is then given by $\widehat{R}_n \triangleq \diag(\hSigma_n)^{-\frac12}\hSigma_n\diag(\hSigma_n)^{-\frac12}$.

Finally, it is straightforward to extend the algorithm for solving
\eqref{eq:robust-SAA-single-expected-value-constraint} to \eqref{eq:robust-SAA-multiple-expected-value-constraint}. The Lagrangian of \eqref{eq:robust-SAA-multiple-expected-value-constraint} is 
\[
\begin{aligned}
L(\theta,\blambda) &\textstyle\triangleq \frac1n\sum_{i=1}^n f(\theta;Z_i) + \sum_{k=1}^K\lambda_k\sup\nolimits_{P:D_\varphi(P\| P_n)\leq\frac{\rho_k}{n}}\bbE_P\big[g_k(\theta;Z)\big] \\
&\textstyle=\frac1n\sum_{i=1}^nf(\theta;Z_i) + \sum_{k=1}^K\lambda_k\textstyle\inf\nolimits_{\mu_k\ge 0,\nu_k\in\bbR}\left\{\frac1n\sum_{i=1}^n\mu_k\varphi^*(\frac{g_k(\theta;Z_i) - \nu_k}{\mu_k}) + \mu_k\frac{\rho_k}{n} + \nu_k\right\},
\end{aligned}
\]
where we recalled the dual form of the robust constraint function \eqref{eq:robust-expected-value-constraint-dual-form} in the second step. We see that evaluating the dual function $\inf_\theta L(\theta,\blambda)$ (at a fixed $\blambda$) entails solving a stochastic optimization problem that is suitable for stochastic approximation. This suggests a similar dual ascent algorithm for solving \eqref{eq:robust-SAA-single-expected-value-constraint}; we skip the details here (see Algorithm \ref{alg:robust-SAA-multiple-expected-value-constraint} in Appendix \ref{sec:dual-ascent-multiple-constraints}).

%% file: sections/simulation.tex
\section{Simulations}
\label{sec:simulation}

We simulate the frequency of constraint satisfaction for the following multi-item newsvendor problem:
\begin{equation}
\label{eq:maximization-problem-multi-item-newsvendor}
\begin{array}{ll}
\max_{\theta\in\Theta} & \mathbb{E}_{P_0} \big[p^\top \min \{Z, \theta\}- c^\top\theta\big] \\
\subjectto & \bbE_{P_0}[(\|Z^{(1)}\|_2^2-\|\theta^{(1)}\|_2^2)_{+}] \leq \varepsilon_1 \\
 & \bbE_{P_0}[(\|Z^{(2)}\|_2^2-\|\theta^{(2)}\|_2^2)_{+}] \leq \varepsilon_2 \\
\end{array}
\end{equation}
where $c \in \bbR^d_{+}$ is the manufacturing cost, $p \in \bbR^d_{+}$ is the sell price, $\theta \in \Theta = [0,100]^d$ is the number of items in stock, $Z\in\bbR^d$ is a random variable with probability distribution $P_0$ representing the demand, and there are $d$ items in total. The distribution $P_0$ is unknown but we observe IID samples $Z_1,\ldots,Z_n$ from $P_0$. All of the items have been partitioned into two groups so that the corresponding demand and stock can be written as $Z = (Z^{(1)}, Z^{(2)})$ and $\theta = (\theta^{(1)}, \theta^{(2)})$. The constraints in the problem exclude stock levels that underestimate the demand too much for each group of items, where $\varepsilon_1,\varepsilon_2 > 0$ indicate tolerance level of such underestimation. The target of the problem is to maximize the profit while satisfying the constraints. It is easy to rewrite the maximization problem \eqref{eq:maximization-problem-multi-item-newsvendor} as a minimization problem with expected value constraints in the form of \eqref{eq:multiple-expected-value-constraint} so that we can apply our method \eqref{eq:robust-SAA-multiple-expected-value-constraint}. We pick $P_0$ as multivariate Gaussian with independent components so that the two constraints are generally uncorrelated with each other (see Appendix \ref{sec:details-simulations} for details).

Throughout the simulations, we solve \eqref{eq:robust-SAA-multiple-expected-value-constraint} with $\brho = (z_\alpha, z_\alpha)^\top$ for $\alpha \in \{0.4, 0.25, 0.1, 0.05, 0.005\}$. As suggested by our asymptotic theory in Section \ref{sec:multiple-expected-value-constraints}, the nominal probability of constraint satisfaction is $1-\alpha$ for each constraint and $(1-\alpha)^2$ for both constraints due to the independence setup.

\begin{figure}[h]
\centering
\begin{minipage}{0.32\textwidth}
  \centering
  \includegraphics[width=1\linewidth]{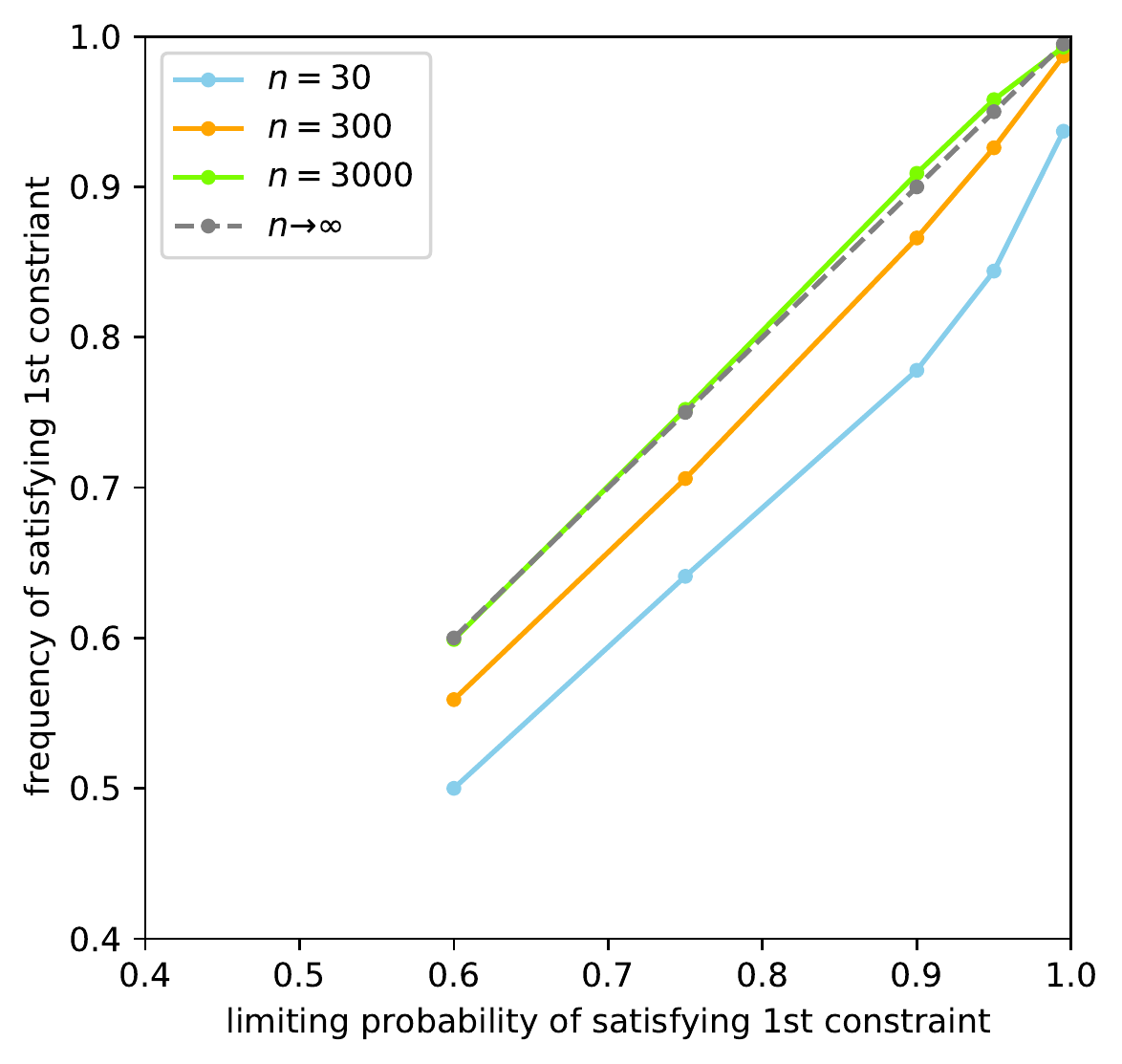}
\end{minipage}
\begin{minipage}{0.32\textwidth}
  \centering
  \includegraphics[width=1\linewidth]{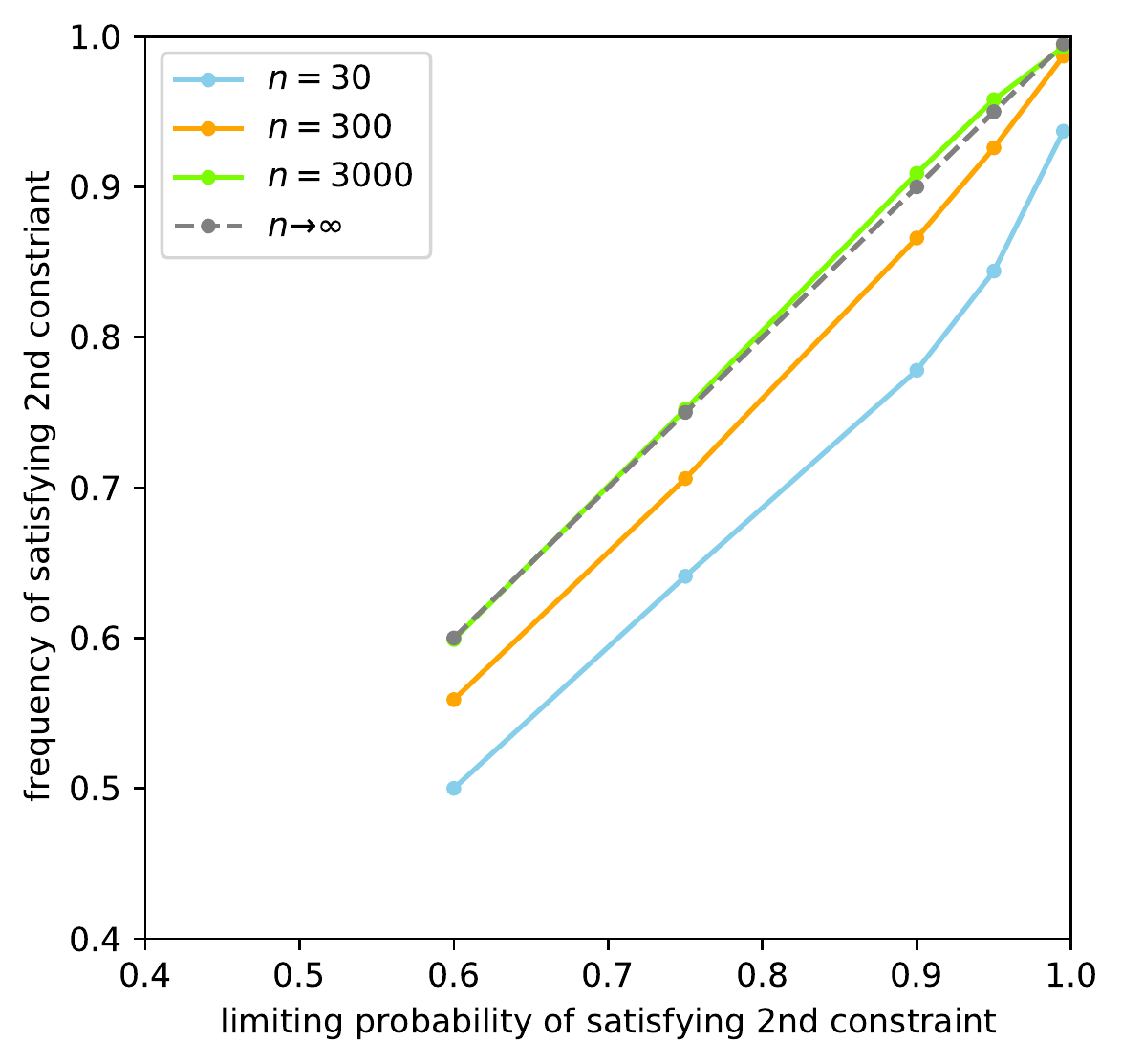}
\end{minipage}
\begin{minipage}{0.32\textwidth}
  \centering
  \includegraphics[width=1\linewidth]{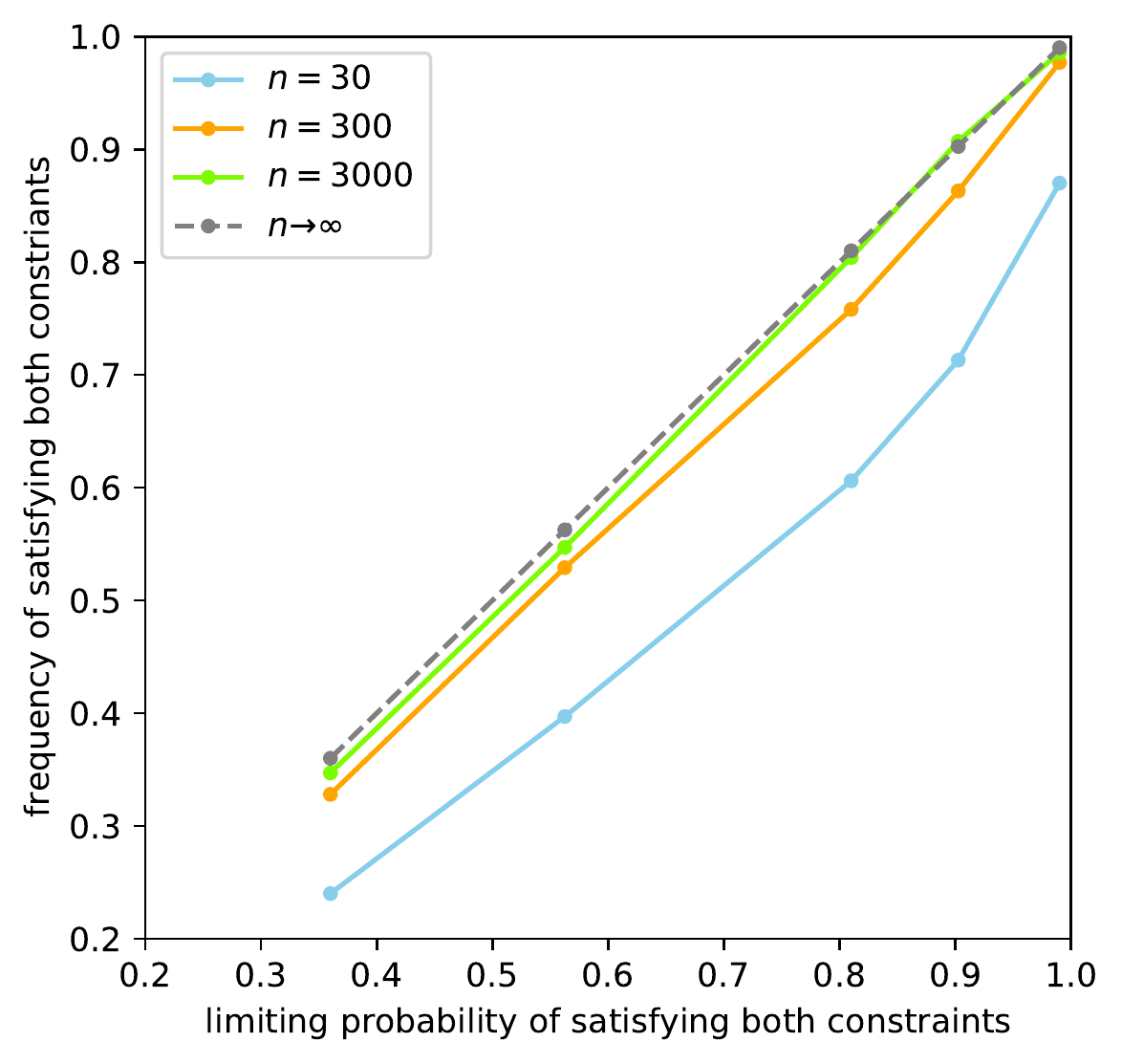}
\end{minipage}
\caption{Frequency versus limiting probability of constraint satisfaction of the first constraint (left), the second constraint (middle), and both of the constraints (right). }
\label{fig:1}
\end{figure}

In Figure \ref{fig:1}, we plot frequencies of constraint satisfaction for each constraint and both constraints, all of which are averaged over $1000$ replicates. As the sample size $n$ grows, the frequency versus probability curve converges to the theoretical dashed line of limiting probability of constraint satisfaction, validating our theory in the large sample regime. For more simulations (\eg, single constraint, two dependent constraints) we refer to Appendix \ref{sec:details-simulations}.

%% file: sections/fairness.tex
\section{Application to fair machine learning}
\label{sec:fairness}

As ML models are deployed in high-stakes decision making and decision support roles, the fairness of the models has come under increased scrutiny. In response, there is a flurry of recent work on mathematical definitions of algorithmic fairness \cite{dwork2011Fairness,hardt2016Equality,kusner2018Counterfactual} and algorithms to enforce the definitions \cite{agarwal2018Reductions,cotter2019TwoPlayer,yurochkin2020SenSeI}. 

A prominent class of fairness definitions is \emph{group fairness}; such definitions require equality of certain metrics (\eg\ false/true positive rates) among demographic groups. For example, consider a fair binary classification problem. Let $\cX\subset\bbR^d$ be the input space, $\{0,1\}$ be the set of possible labels, and $\cA$ be the set of possible values of the protected/sensitive attribute. In this setup, training and test examples are tuples of the form $(X,A,Y)\in\cX\times\cA\times\cY$, and a classifier is a map $f:\cX\to\{0,1\}$.
A popular definition of algorithmic fairness for binary classification is \emph{equality of opportunity} \cite{hardt2016Equality}. 

\begin{definition}[equality of opportunity]
\label{def:equal-opportunity}
Let $Y=1$ be the advantaged label that is associated with a positive outcome and $\hY \triangleq f(X)$ be the output of the classifier. Equality of opportunity entails $\bbP\{\hY=1\mid A=a,Y=1\} = \bbP\{\hY=1\mid A=a^\prime,Y=1\}$ for all $a,a^\prime\in\cA$.
\end{definition}
Equality of opportunity, or true positive rate parity, means that the prediction $\widehat{Y} = h(X)$ conditioned on the advantaged label $Y=1$ is statistically independent of the protected attribute $A$.
Furthermore, an approximate version of equality of opportunity can be readily defined. We say that $\widehat{Y} = h(X)$ satisfies \emph{$\varepsilon$-equality of opportunity} if $\bbP\{\hY=1\mid A=a,Y=1\} - \bbP\{\hY=1\mid A=a^\prime,Y=1\} \leq \varepsilon$ for for all $a,a'\in\cA$. In this case, $\varepsilon>0$ represents a practitioner's \emph{tolerance} for fairness violations.

Given a parametric model space $\cH = \{f_\theta(\cdot): \theta \in \Theta\}$ and loss function $\ell: \Theta\times\cX\times\cY \to \bbR_+$, an in-processing fair ML routine is to minimize the (empirical) risk $\bbE \left[\ell(\theta; X, Y)\right]$ while satisfying some fairness constraints. Most commonly, definitions of group fairness (including equality of opportunity, demographic parity, and more) can be written as a special example of a general set of linear constraints \cite{agarwal2018Reductions, agarwal2019Fair} of the form $\bM\boldsymbol{\mu}(\theta)\leq\bc$, where matrix $\bM\in\bbR^{K \times T}$ and vector $\bc \in \bbR^{K}$ encode the constraints; $\boldsymbol{\mu}(\theta): \Theta \to \bbR^T$ is a vector of (conditional) moments
$\mu_t(\theta) = \bbE\left[h_t(X, A, Y, \theta)\mid \cE_t\right]$ for $t\in[T]$; $g_t:\cX\times\cA\times\cY\times\Theta\to\bbR$; event $\cE_t$ is defined with respect to $(X, A, Y)$.

This framework fits to our methodology if we note that each (conditional) moment can be written as
\begin{equation}\label{eq:ratio-of-expected-values}
    \mu_t(\theta) = \frac{\bbE_{(X,A,Y)\sim P_0}\big[h_t(X,A,Y,\theta)\times \ones\left\{\cE_t(X,Y,A)\right\}\big]}{\bbE_{(X,A,Y)\sim P_0}\big[ \ones\left\{\cE_t(X,Y,A)\right\}\big]}.
\end{equation}
Here the indicator $\ones\left\{\cE_t\right\}$ takes value $1$ if the event $\cE_t$ happens, and $0$ otherwise. Moreover, we use $\cE_t(X, A, Y)$ to emphasize that $\cE_t$ only depends on $(X, Y, A)$ but not on $\theta$ in any way.

Note that \eqref{eq:ratio-of-expected-values} is a ratio of expected values, which is a non-linear statistical functional of $P_0$. To use our method, we first replace the denominator of $\mu_t(\theta)$ with an estimator, such as the unbiased estimator $\widehat{\bbP}(\cE_t) = \frac{1}{n}\sum_{i=1}^n \ones\left\{\cE_t(X_i,A_i,Y_i)\right\}$. The resulting plug-in estimation of $\mu_t(\theta)$ then becomes linear in $P_0$, allowing us to apply our method (see similar tricks in \cite{cotter2019training}).
We describe the application of our method to $\varepsilon$-equality of opportunity in Appendix \ref{sec:instantiate-EO}.

\subsection{A two-stage method for unknown active set}
\label{sec:unknown-active-set}

In practice, it is probable that only a subset of the constraints are active. Furthermore, we do not know beforehand whether or not a constraint is active in the true population problem. To handle this scenario, we propose a two-stage method:
\begin{enumerate}
    \item At the first stage, we solve the sample average approximation (SAA) problem \eqref{eq:robust-SAA-multiple-expected-value-constraint} with $\brho = \zeros_K$. By doing so, we identify the active set of the SAA problem.
    \item At the second stage, we solve \eqref{eq:robust-SAA-multiple-expected-value-constraint} with $\brho$ such that $\rho_k$ is a positive number only if the $k$-th constraint, $k \in [K]$, was identified as active at the first stage.
\end{enumerate}
In Appendix \ref{sec:details-unknown-active-set}, we show that the two-stage method also enjoys the calibration property (similar to Theorem \ref{thm:robust-SAA-multiple-expected-value-constraint-asymptotically-exact-constraint-satisfaction} and Corollary \ref{cor:robust-SAA-multiple-expected-value-constraint-asymptotically-exact-constraint-satisfaction}) under standard assumptions (\ie, strict complementarity). At a high level, the limiting probability of satisfying the true constraints depends solely on the correlation structure between active constraints and the uncertainty set radii for active constraints, as long as the SAA problem identifies active constraints with probability tending to $1$.

\subsection{Proxy dual function for non-differentiable constraints}
Constraint functions in fair ML are often non-differentiable. For instance, fairness metrics are typically linear combinations of indicators that result in non-differentiable rate constraints \citep{cotter2019training,cotter2019Optimization,cotter2019TwoPlayer}.
This prevents the use of any gradient-based optimization algorithms.
Fortunately, only the dual function evaluation step in Algorithm \ref{alg:robust-SAA-single-expected-value-constraint} requires access to gradients.
Therefore, we can modify the algorithm by: (1) introducing proxy dual function, which uses a differentiable surrogate $\Tilde{g}$ instead of the non-differentiable $g$ in the dual function evaluation step; (2) keeping $g$ in the dual ascent step. For an indicator function $h(t) = \ones\{t > 0\}$, one can replace it by sigmoidal function $h_1(t) = (1 + e^{-at})^{-1}$ or hinge upper bound $h_2(t) = \max\{0, t+1\}$ to produce smooth surrogates for non-differentiable rate constraints \citep{davenport2010tuning, eban2017scalable, cotter2019Optimization}. We summarize the proxy dual ascent algorithm in Appendix \ref{sec:proxy-dual-ascent-algorithm}.

\subsection{Adult experiments}
\label{sec:adult}
We compare the frequency of constraint satisfaction (at test time) of the sample average approximation and our methods with nominal probability $0.60, 0.75, 0.90, 0.95$ using the Adult dataset from UCI \citep{Dua:2019}. The classification task is to predict whether an individual's income per year is higher than $\$50$K. The fairness goal is $\varepsilon$-demographic parity ($\varepsilon$-DP): $|\bbP(\hY = 1\mid A = 1) - \bbP(\hY = 1\mid A = 0)|\leq \varepsilon$, where $A = 1$ for male is the advantaged group and $A = 0$ for female is the disadvantaged group. We use a logistic regression model for classification and techniques in this section for implementation.

\begin{figure}[h]
\centering
\begin{minipage}{0.48\textwidth}
  \centering
  \includegraphics[width=1\linewidth]{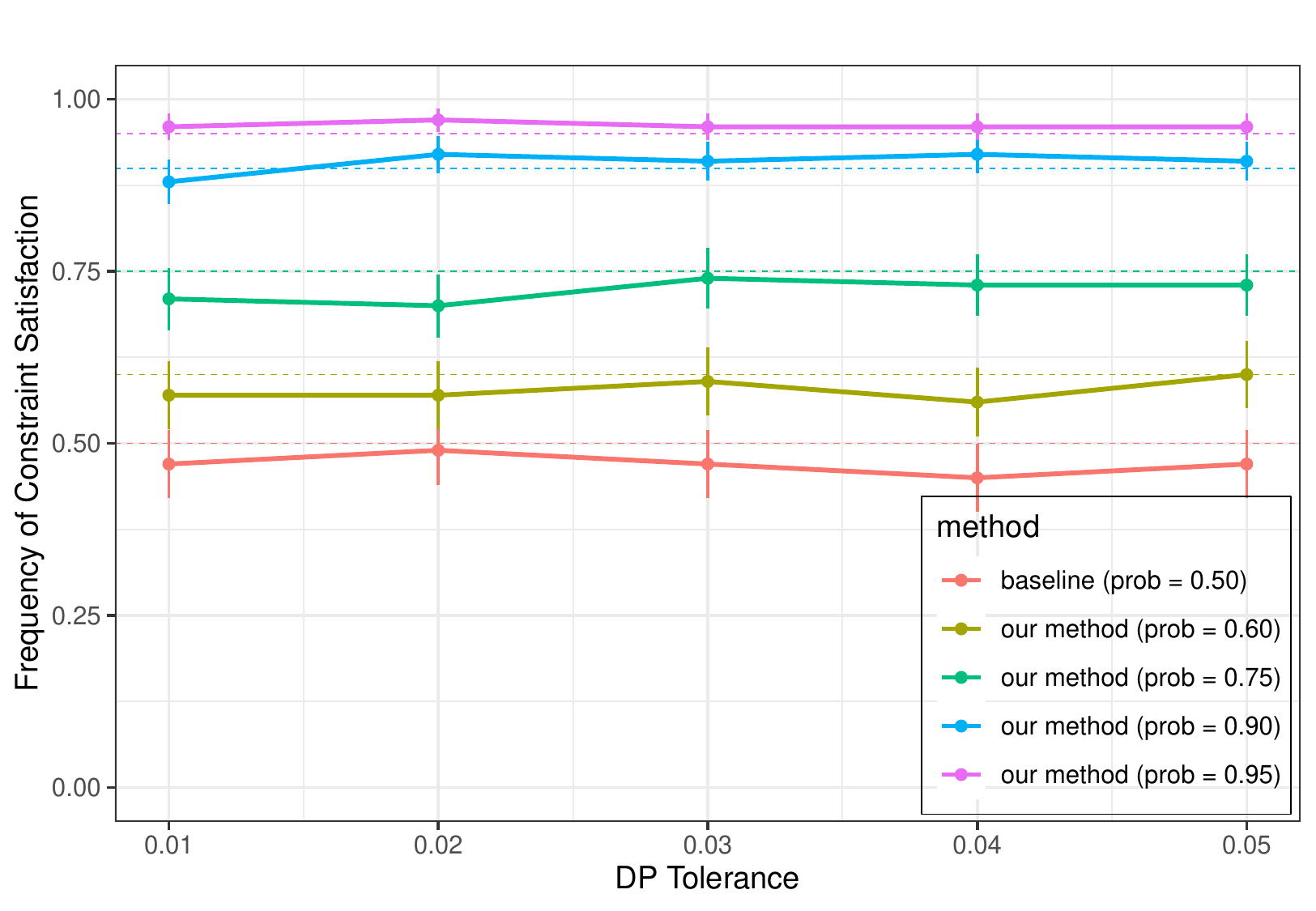}
\end{minipage}
\begin{minipage}{0.48\textwidth}
  \centering
  \includegraphics[width=1\linewidth]{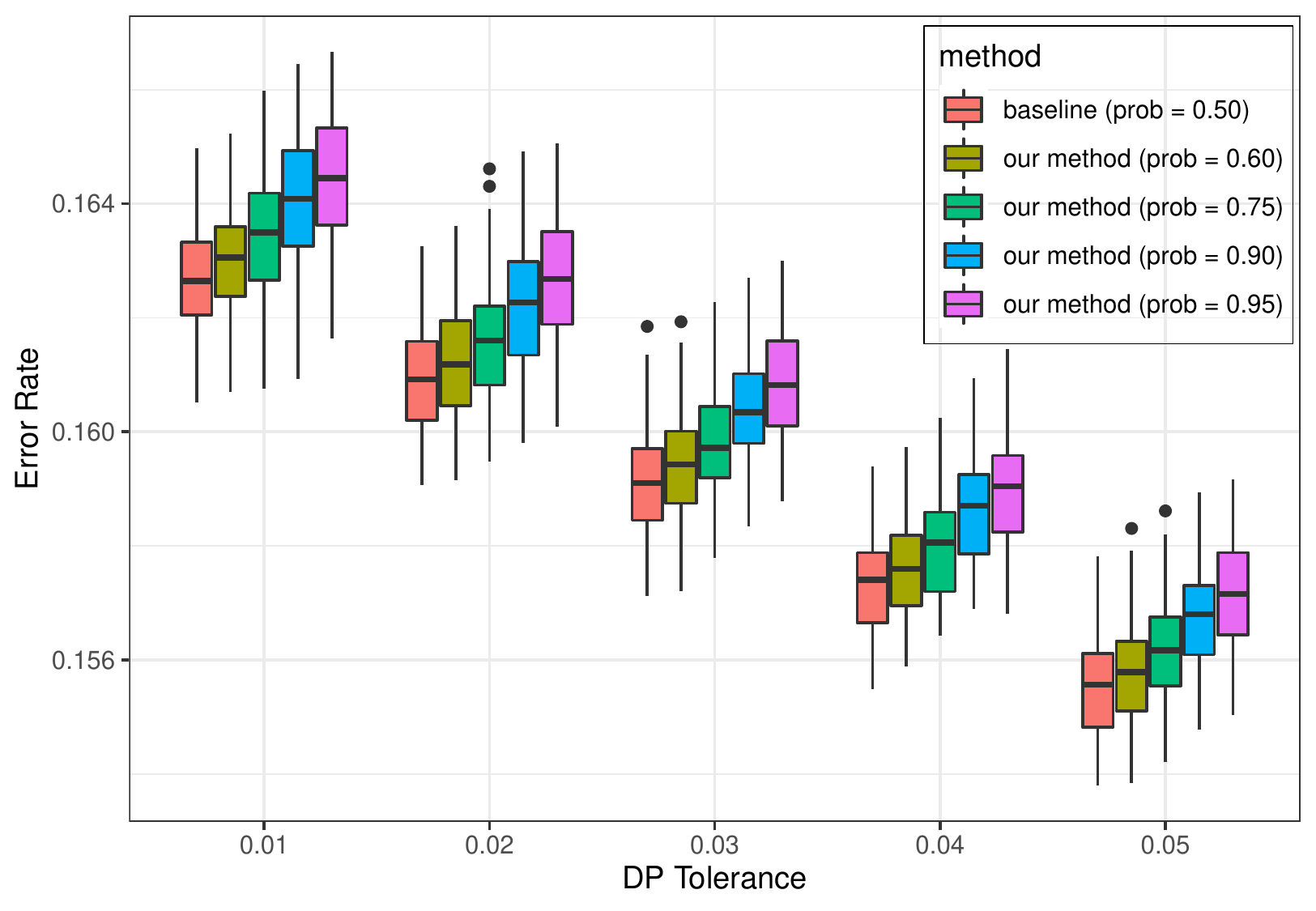}
\end{minipage}
\caption{Frequency of constraint satisfaction (left) and classification error rate (right) for different demographic parity tolerance $\varepsilon \in \{0.01, 0.02, 0.03, 0.04, 0.05\}$. Baseline (sample average approximation, SAA) and our methods (with nominal probability $0.60, 0.75, 0.90,0.95$) are compared.}
\label{fig:2}
\end{figure}

In Figure \ref{fig:2}, we have line plots for frequency of constraint satisfaction and box plots for classification error rate, all of which are summarized over $100$ replicates. The left panel shows that solving \eqref{eq:multiple-expected-value-constraint} directly leads to one half chance of constraint violation, while our method's constraint satisfaction frequency matches its nominal value. The price of a higher chance of test-time fairness satisfaction is an increase in classification error rate as shown in the right panel. From the baseline to $95\%$ chance of fairness satisfaction, we basically trade off $2\%$ increase in error rate. We refer to Appendix \ref{sec:details-experiments} and \ref{sec:more-baseline} for details and more experiments.

%% file: sections/summary.tex
\section{Summary and discussion}
\label{sec:summary}
We explore the problem of exact constraint satisfaction probability in stochastic optimization with expected-value constraints.
We propose a distributionally robust reformulation of data-dependent constraints and provide a theoretical guarantee of constraint satisfaction with an asymptotically exact probability specified by the user.
For solving the reformulated problem, a scalable dual ascent algorithm and its variants are proposed.
The computational cost of our algorithm is comparable to that of a standard distributionally robust optimization problem.
Our theory on exact constraint satisfaction probability is validated via simulations on the resource-constrained newsvendor problem.
The efficacy of our methods is empirically demonstrated on fair machine learning applications.

Some data-dependent constraints are by nature \emph{non-linear} in the underlying probability measure.
For example, \eqref{eq:ratio-of-expected-values} is a ratio of expected values.
An intriguing direction for future research is to generalize the methods and theory developed in this work to constraints on non-linear functions of expected values.
Such forms of constraints are known as \emph{statistical functionals} in statistics literature \citep{fernholz2012mises}.
The non-linear dependence of the constraint function on the probability measure precludes the stochastic approximation as a general way of evaluating the dual function, as the constraint function no longer admits a dual form \eqref{eq:robust-expected-value-constraint-dual-form}, calling for the development of a new algorithm.

%% file: sections/appendix.tex
\section{Proofs of Theorem \ref{thm:robust-SAA-single-expected-value-constraint-asymptotically-exact-constraint-satisfaction} and Corollary \ref{cor:robust-SAA-single-expected-value-constraint-asymptotically-exact-constraint-satisfaction}}
\label{sec:proofs-for-section-2}

Note that Theorem \ref{thm:robust-SAA-multiple-expected-value-constraint-asymptotically-exact-constraint-satisfaction} implies Theorem \ref{thm:robust-SAA-single-expected-value-constraint-asymptotically-exact-constraint-satisfaction} and Corollary \ref{cor:robust-SAA-multiple-expected-value-constraint-asymptotically-exact-constraint-satisfaction} implies Corollary \ref{cor:robust-SAA-single-expected-value-constraint-asymptotically-exact-constraint-satisfaction} by letting $K = 1$. Therefore, it is sufficient to prove Theorem \ref{thm:robust-SAA-multiple-expected-value-constraint-asymptotically-exact-constraint-satisfaction} and Corollary \ref{cor:robust-SAA-multiple-expected-value-constraint-asymptotically-exact-constraint-satisfaction}, whose proofs can be found in Appendix \ref{sec:proof-thm-3.1} and \ref{sec:proof-cor-3.2} respectively. \hfill $\square$

\section{Proof of Theorem \ref{thm:robust-SAA-multiple-expected-value-constraint-asymptotically-exact-constraint-satisfaction}}
\label{sec:proof-thm-3.1}

Consider a stochastic optimization problem with $K$ expected value constraints
\begin{equation*}\label{eq:problem-population}
    (\cP_0):\quad\theta^\star \in \argmin_{\theta \in \Theta} \left\{ \bbE_{Z\sim P_0} \left[f(\theta; Z)\right]: \bbE_{Z\sim P_0} \left[g_k(\theta; Z)\right]\leq 0, k\in[K]\right\}.
\end{equation*}

Our proposed robust constraint method solves
\begin{equation*}\label{eq:problem-dro}
    (\cP_n):\quad\htheta_{n,\brho} \in \argmin_{\theta \in \Theta} \left\{ \bbE_{Z\sim P_n} \left[f(\theta; Z)\right]: \sup_{D_\varphi (Q\|P_n)\leq \rho_k/n} \bbE_{Z\sim Q} \left[g_k(\theta; Z)\right]\leq 0, k\in[K]\right\},
\end{equation*}
where $\brho = (\rho_1,\ldots,\rho_K)^\top$ is the collection of critical radii of uncertainty sets. Here we denote the empirical distribution $\hP_n$ by $P_n$ for notation simplicity.

As a special case of our robust method, the sample average approximation (SAA) or empirical risk minimization (ERM) solves
\begin{equation*}\label{eq:problem-saa}
    \htheta_{n,\zeros_K} \in \argmin_{\theta \in \Theta} \left\{ \bbE_{Z\sim P_n} \left[f(\theta; Z)\right]: \bbE_{Z\sim P_n} \left[g_k(\theta; Z)\right]\leq 0, k\in[K]\right\}.
\end{equation*}

We denote
\begin{equation*}
    F(\theta) = \bbE_{Z\sim P_0} \left[f(\theta; Z)\right]\quad\text{and}\quad \widehat{F}_n(\theta) = \bbE_{Z\sim P_n} \left[f(\theta; Z)\right],
\end{equation*}
\begin{equation*}
    G_k(\theta) = \bbE_{Z\sim P_n} \left[g_k(\theta; Z)\right]\quad\text{and}\quad \widehat{G}_{kn}(\theta) = \sup_{D_\varphi (Q\|P_n)\leq \rho_k/n} \bbE_{Z\sim Q} \left[g_k(\theta; Z)\right]\quad\text{for}\quad k\in[K],
\end{equation*}
and
\begin{equation*}
    \bG(\theta) = \left(\begin{array}{c}
        G_1(\theta)  \\
        \vdots \\
        G_K(\theta)
    \end{array}\right)\quad\text{and}\quad
    \wbG_n(\theta) = \left(\begin{array}{c}
        \wG_{1n}(\theta)  \\
        \vdots \\
        \wG_{Kn}(\theta)
    \end{array}\right).
\end{equation*}

Note that $\wF_n(\cdot)$ and $\wG_{kn}(\cdot)$'s are random functions serving as approximations to $F(\cdot)$ and $G_k(\cdot)$'s. Consider the Lagrangian functions
\begin{equation*}
    L(\theta, \blambda) = F(\theta) + \blambda^\top \bG(\theta) = F(\theta) + \sum_{k=1}^K \lambda_k G_k(\theta)
\end{equation*}
and
\begin{equation*}
    \wL_n(\theta, \blambda) = \wF_n(\theta) + \blambda^\top \wbG_n(\theta) = \wF_n(\theta) + \sum_{k=1}^K \lambda_k \wG_{kn}(\theta)
\end{equation*}
of the programs $(\cP_0)$ and $(\cP_n)$ respectively.

\begin{lemma}[Theorem 6.6.2 in \cite{rubinstein1993discrete}]\label{thm:RS-6.6.2}
Suppose that:
\begin{enumerate}[label=(\roman*)]
    \item The functions $F(\theta)$ and $G_k(\theta)$, $k\in[K]$, are twice continuously differentiable.
    \item The true program $(\cP_0)$ has a unique optimal solution $\theta^\star$ and a unique vector $\blambda^\star$ of the Lagrange multipliers with $\theta^\star$ being an interior point of $\Theta$.
    \item The Hessian matrix $\nabla^2 L(\theta^\star, \blambda^\star)$ is positive definite.
    \item The random functions $\wG_{kn}(\theta)$,$k\in[K]$, are Lipschitz continuous in a neighborhood of $\theta^\star$ and differentiable at $\theta^\star$ with probability 1.
    \item \begin{equation*}
        \|\Delta_{in}(\theta^\star)\|_2 = O_p(n^{-1/2}),\quad i = 1,2,3
    \end{equation*}
    and there is a neighborhood $U$ of $\theta^\star$ such that
    \begin{equation*}
        \sup_{\theta\in U} \frac{\|\Delta_{in}(\theta) - \Delta_{in}(\theta^\star)\|_2}{n^{-1/2} + \|\theta - \theta^\star\|_2} = o_p(1),\quad i = 1,2,3.
    \end{equation*}
    Here we define random mappings $\Delta_{1n}(\theta) = \nabla \wF_n(\theta) - \nabla F(\theta)$, $\Delta_{2n}(\theta) = \wbG_n(\theta) - \bG(\theta)$, and $\Delta_{3n}(\theta) = \nabla \wbG_n(\theta) - \nabla \bG(\theta)$.
    \item Random vectors $\sqrt{n}(\nabla \wL_n(\theta^\star, \blambda^\star), \wbG_n(\theta^\star))$ converge in distribution as $n\to\infty$ to a random vector $\bY = (\bY_1, \bY_2)$.
\end{enumerate}
Let $\htheta_n$ be an optimal solution of $(\cP_n)$ converging in probability as $n\to\infty$ to $\theta^\star$. Then
\begin{equation*}
    \sqrt{n}(\htheta_n - \theta^\star) \overset{d}{\longrightarrow} \bar{\bx}(\bY)
\end{equation*}
where $\bar{\bx} = \bar{\bx}(\bY)$ is the optimal solution to the quadratic programming problem
\begin{equation*}
    \begin{array}{ll}
    \underset{\boldsymbol{x}}{\operatorname{minimize}} & \bx^\top \bY_1 + \frac{1}{2} \bx^\top \nabla^2 L(\theta^\star, \blambda^\star) \bx  \\
    \operatorname{subject~to} &  \nabla \bG(\theta^\star)^\top \bx + \bY_2 = \zeros
\end{array}.
\end{equation*}
\end{lemma}

From now on, we use $\varphi(t) = (t-1)^2$, which gives the $\chi^2$-divergence. Lemma \ref{thm:RS-6.6.2} is adapted from Theorem 6.6.2 in \cite{rubinstein1993discrete} under the strict complementarity assumption. Recall the standing assumptions, \emph{(i)}, \emph{(iv)}, \emph{(v)} are guaranteed by the smoothness and concentration assumption, \emph{(ii)} is postulated by the uniqueness assumption, and \emph{(iii)} is ensured by the positive definiteness assumption. Now we derive the limiting distribution required in \emph{(vi)}.

Let $\bbB$ denote the $\ell_2$-ball of radius $1$ in $\bbR^d$. According to Lemma 24 in \cite{duchi2016Variancebased}, for each $k\in[K]$ there exists $\epsilon_k > 0$ such that, with probability 1, there exists an $N_k$ such that $n\geq N_k$ implies
\begin{equation*}
    \wG_{kn}(\theta) = \bbE_{P_n}[g_k(\theta; Z)] + \sqrt{\frac{\rho_k \Var_{P_n}[g_k(\theta;Z)]}{n}}~\text{for all}~\theta \in \theta^\star + \epsilon_k \bbB.
\end{equation*}

Taking $\epsilon_0 = \min\{\epsilon_k:k\in[K]\}$ and $N_0 = \max\{N_k:k\in[K]\}$, we have the following uniform expansion holds, that is,
\begin{equation*}
    \wG_{kn}(\theta) = \bbE_{P_n}[g_k(\theta; Z)] + \sqrt{\frac{\rho_k}{n}\Var_{P_n}[g_k(\theta;Z)]}~\text{for all}~k\in[K]~\text{and}~\theta \in \theta^\star + \epsilon_0 \bbB
\end{equation*}
$P_0$-almost surely given $n\geq N_0$.

Therefore, for sufficiently large $n$, for $k \in [K]$ we have
\begin{align*}
    \wG_{kn}(\theta^\star) &= \bbE_{P_n}[g_k(\theta^\star; Z)] + \sqrt{\frac{\rho_k}{n}\Var_{P_n}[g_k(\theta;Z)]}\\
    &= \bbE_{P}[g_k(\theta^\star; Z)] + \{\bbE_{P_n}[g_k(\theta^\star; Z)] - \bbE_{P_0}[g_k(\theta^\star; Z)]\} + \sqrt{\frac{\rho_k}{n}\left(\Var_{P_0}[g_k(\theta^\star;Z)]+ o_P(1)\right)} \\
    &= G_k(\theta^\star) + \{\bbE_{P_n}[g_k(\theta^\star; Z)] - \bbE_{P_0}[g_k(\theta^\star; Z)]\} + \sqrt{\frac{\rho_k}{n}\Var_{P_0}[g_k(\theta^\star;Z)]} + o_P(n^{-1/2})
\end{align*}
and
\begin{align*}
    \nabla \wG_{kn}(\theta^\star) =& \bbE_{P_n}[\nabla g_k(\theta^\star; Z)] + \nabla \sqrt{\frac{\rho_k}{n}\Var_{P_n}[g_k(\theta;Z)]}\\
    =& \bbE_{P}[\nabla g_k(\theta^\star; Z)] + \{\bbE_{P_n}[\nabla g_k(\theta^\star; Z)] - \bbE_{P_0}[\nabla g_k(\theta^\star; Z)]\} + \\
    &\sqrt{\frac{\rho_k}{n}} \frac{\mathbb{E}_{P_n}[\left(\nabla g_k(\theta^\star, X)-\mathbb{E}_{P_n}[\nabla g_k(\theta^\star;Z)]\right)\left(g_k(\theta^\star;Z)-\mathbb{E}_{P_n}[g_k(\theta^\star;Z)]\right)]}{\sqrt{\Var_{P_n}[g_k(\theta^\star;Z)]}}\\
    =& \nabla \bbE_{P_0}[g_k(\theta^\star; Z)] + \{\bbE_{P_n}[\nabla g_k(\theta^\star; Z)] - \bbE_{P_0}[\nabla g_k(\theta^\star; Z)]\} + \\
    &\sqrt{\frac{\rho_k}{n}}\left(\frac{\Cov_{P_0}(\nabla g_k(\theta^\star;Z), g_k(\theta^\star;Z))}{\sqrt{\Var_{P_0}[g_k(\theta^\star;Z)]}} + o_P(1)\right) \\
    =& \nabla G_k(\theta^\star) + \{\bbE_{P_n}[\nabla g_k(\theta^\star; Z)] - \bbE_{P_0}[\nabla g_k(\theta^\star; Z)]\} +\\
    &\sqrt{\frac{\rho_k}{n}} \frac{\Cov_{P_0}(\nabla g_k(\theta^\star;Z), g_k(\theta^\star;Z))}{\sqrt{\Var_{P_0}[g_k(\theta^\star;Z)]}} + o_P(n^{-1/2}).
\end{align*}

For the objective function and its empirical counterpart, we have
\begin{equation*}
    \wF(\theta^\star) = \bbE_{P_n}[f(\theta^\star; Z)] = F(\theta^\star) + \{\bbE_{P_n}[f(\theta^\star; Z)] - \bbE_{P_0}[f(\theta^\star; Z)]\}
\end{equation*}
and
\begin{equation*}
    \nabla\wF(\theta^\star) = \bbE_{P_n}[\nabla f(\theta^\star; Z)] = \nabla F(\theta^\star) + \{\bbE_{P_n}[\nabla f(\theta^\star; Z)] - \bbE_{P_0}[\nabla f(\theta^\star; Z)]\}.
\end{equation*}

Now we derive the the limiting distribution of random vectors $\sqrt{n}(\nabla \wL_n(\theta^\star, \blambda^\star), \wbG_n(\theta^\star))$. For simplicity of notations, we denote $P_n m = \bbE_{P_n} [m (\theta^\star;Z)]$ and $P_0 m = \bbE_{P_0} [m(\theta^\star;Z)]$ for any random function $m(\theta; Z)$, $ \Cov(\nabla g_k, g_k) = \Cov_{P_0}(\nabla g_k(\theta^\star;Z), g_k(\theta^\star;Z))$ and $\Var[g_k] = \Var_{P_0}[g_k(\theta^\star;Z)]$. Then we have
\begin{align*}
    &\begin{bmatrix}
    \nabla \wL_n(\theta^\star, \blambda^\star)\\
    \wbG_n(\theta^\star)
    \end{bmatrix} \\
    =& \begin{bmatrix}
    \nabla\wF_n(\theta^\star) + \sum_{k=1}^K \lambda_k^\star \nabla\wG_{kn}(\theta^\star)\\
    \vdots \\
    \wG_{kn}(\theta^\star) \\
    \vdots
    \end{bmatrix} \\
    =& \begin{bmatrix}
    \underbrace{\nabla F(\theta^\star) + \sum_{k=1}^K \lambda_k^\star \nabla G_{k}(\theta^\star)}_{=0~\text{due to KKT condition}} + (P_n\nabla f - P_0\nabla f) + \sum_{k=1}^K \lambda_k^\star (P_n\nabla g_k - P_0\nabla g_k) + \frac{1}{\sqrt{n}}\sum_{k=1}^K \frac{\lambda_k^\star \sqrt{\rho_k}\Cov(\nabla g_k, g_k)}{\sqrt{\Var[g_k]}} \\
    \vdots \\
    \underbrace{G_k(\theta^\star)}_{=0~\text{due to active constraint}} + (P_n g_k - P_0 g_k) + \sqrt{\frac{\rho_k}{n}\Var[g_k]}\\
    \vdots
    \end{bmatrix} \\
    &+ o_P\left(\frac{1}{\sqrt{n}}\right) \\
    =& \frac{1}{\sqrt{n}} \begin{bmatrix}
    \sum_{k=1}^K \frac{\lambda_k^\star \sqrt{\rho_k}\Cov(\nabla g_k, g_k)}{\sqrt{\Var[g_k]}} \\
    \vdots \\
    \sqrt{\rho_k \Var[g_k]}\\
    \vdots
    \end{bmatrix} + \left\{P_n\begin{bmatrix}
    \nabla f + \sum_{k=1}^K \lambda_k^\star \nabla g_k \\
    \vdots \\
     g_k\\
    \vdots
    \end{bmatrix} - P_0\begin{bmatrix}
    \nabla f + \sum_{k=1}^K \lambda_k^\star \nabla g_k \\
    \vdots \\
     g_k\\
    \vdots
    \end{bmatrix}\right\} + o_P\left(\frac{1}{\sqrt{n}}\right).
\end{align*}
By central limit theorem,
\begin{equation*}
    \sqrt{n}\left\{P_n\begin{bmatrix}
    \nabla f + \sum_{k=1}^K \lambda_k^\star \nabla g_k \\
    \vdots \\
     g_k\\
    \vdots
    \end{bmatrix} - P_0\begin{bmatrix}
    \nabla f + \sum_{k=1}^K \lambda_k^\star \nabla g_k \\
    \vdots \\
     g_k\\
    \vdots
    \end{bmatrix}\right\} \overset{d}{\longrightarrow} \cN\left(\zeros, \begin{bmatrix}
    \Sigma_{11} & \Sigma_{12} \\
    \Sigma_{21} & \Sigma_{22}
    \end{bmatrix}\right),
\end{equation*}
where
\begin{align*}
    \Sigma_{11} &= \Var_{P_0}\left[\nabla f(\theta^\star;Z) + \sum_{k=1}^K \lambda_k^\star \nabla g_k(\theta^\star;Z)\right] \in \bbR^{d\times d}, \\
    \Sigma_{12} &= \Cov_{P_0}\left(\nabla f(\theta^\star;Z) + \sum_{k=1}^K \lambda_k^\star \nabla g_k(\theta^\star;Z), \bG(\theta^\star;Z)\right) \in \bbR^{d\times K}, \\
    \Sigma_{21} &= \Sigma_{12}^\top, \\
    \Sigma_{22} &= \Var_{P_0} [\bG(\theta^\star;Z)] \in \bbR^{K \times K}.
\end{align*}

By Slutsky's theorem, 
\begin{align*}
    &\sqrt{n} \begin{bmatrix}
    \nabla \wL_n(\theta^\star, \blambda^\star)\\
    \wbG_n(\theta^\star)
    \end{bmatrix} \\
    =& \begin{bmatrix}
    \sum_{k=1}^K \frac{\lambda_k^\star \sqrt{\rho_k}\Cov(\nabla g_k, g_k)}{\sqrt{\Var[g_k]}} \\
    \vdots \\
    \sqrt{\rho_k \Var[g_k]}\\
    \vdots
    \end{bmatrix} + \sqrt{n}\left\{P_n\begin{bmatrix}
    \nabla f + \sum_{k=1}^K \lambda_k^\star \nabla g_k \\
    \vdots \\
     g_k\\
    \vdots
    \end{bmatrix} - P_0\begin{bmatrix}
    \nabla f + \sum_{k=1}^K \lambda_k^\star \nabla g_k \\
    \vdots \\
     g_k\\
    \vdots
    \end{bmatrix}\right\} + o_P(1) \\
    \dto& \cN\left(\begin{bmatrix}
    \mu_1 \\
    \mu_2
    \end{bmatrix}, \begin{bmatrix}
    \Sigma_{11} & \Sigma_{12} \\
    \Sigma_{21} & \Sigma_{22}
    \end{bmatrix}\right),
\end{align*}
where
\begin{align*}
    \mu_1 &= \sum_{k=1}^K \frac{\lambda_k^\star \sqrt{\rho_k} \Cov_{P_0}(\nabla g_k(\theta^\star;Z), g_k(\theta^\star;Z))}{\sqrt{\Var_{P_0}[g_k(\theta^\star;Z)]}} \in \bbR^{d},\\
    \mu_2 &= \begin{bmatrix}
    \sqrt{\rho_1\Var_{P_0}[g_1(\theta^\star;Z)]} \\
    \vdots \\
    \sqrt{\rho_K\Var_{P_0}[g_K(\theta^\star;Z)]} 
    \end{bmatrix} \in \bbR^{K}.
\end{align*}
Therefore, we conclude that the limiting distribution of $\sqrt{n}(\nabla \wL_n(\theta^\star, \blambda^\star), \wbG_n(\theta^\star))$ is
\begin{equation*}
    (\bY_1, \bY_2) \sim \cN\left(\begin{bmatrix}
    \mu_1 \\
    \mu_2
    \end{bmatrix}, \begin{bmatrix}
    \Sigma_{11} & \Sigma_{12} \\
    \Sigma_{21} & \Sigma_{22}
    \end{bmatrix}\right).
\end{equation*}
By Lemma \eqref{thm:RS-6.6.2}, we have
\begin{equation*}
    \sqrt{n}(\htheta_n - \theta^\star) \overset{d}{\longrightarrow} \bar{\bx},
\end{equation*}
where $\bar{\bx}$ is given by the linear system
\begin{equation*}
    \underbrace{\begin{bmatrix}
    \nabla^2 L(\theta^\star, \blambda^\star) & \nabla \bG(\theta^\star)\\
    \nabla \bG(\theta^\star)^\top & \zeros
    \end{bmatrix}}_{\triangleq B} \begin{bmatrix}
    \bar{\bx}\\
    \bar{\blambda}
    \end{bmatrix} = - \begin{bmatrix}
    \bY_1\\
    \bY_2
    \end{bmatrix} \sim \cN\left(- \begin{bmatrix}
    \mu_1 \\
    \mu_2
    \end{bmatrix}, \begin{bmatrix}
    \Sigma_{11} & \Sigma_{12} \\
    \Sigma_{21} & \Sigma_{22}
    \end{bmatrix}\right),
\end{equation*}
or
\begin{equation}\label{eq:solve-linear-system}
    \begin{bmatrix}
    \bar{\bx}\\
    \bar{\blambda}
    \end{bmatrix} \sim \cN\left(- B^{-1}\begin{bmatrix}
    \mu_1 \\
    \mu_2
    \end{bmatrix}, B^{-1} \begin{bmatrix}
    \Sigma_{11} & \Sigma_{12} \\
    \Sigma_{21} & \Sigma_{22}
    \end{bmatrix} B^{-1}\right),
\end{equation}
which implies $\sqrt{n}(\htheta_n - \theta^\star) \overset{d}{\longrightarrow} \bar{\bx} \sim \cN(\bar{\mu}, \bar{\Sigma})$ for some $\bar{\mu}$ and $\bar{\Sigma}$ determined by \eqref{eq:solve-linear-system}.

By delta method, we have
\begin{equation*}
    \sqrt{n}
    \begin{bmatrix}
        \bbE_{P_0}[g_1(\htheta_n;Z)]\\
        \vdots \\
        \bbE_{P_0}[g_K(\htheta_n;Z)]
    \end{bmatrix} = \sqrt{n}\bG(\htheta_n) = \sqrt{n}\{\bG(\htheta_n) - \underbrace{\bG(\theta^\star)}_{=0}\} \overset{d}{\longrightarrow} \cN(\nabla \bG(\theta^\star)^\top \bar{\mu},  \nabla\bG(\theta^\star)^\top \bar{\Sigma}  \nabla\bG(\theta^\star)).
\end{equation*}

Now we calculate $\nabla \bG(\theta^\star)^\top \bar{\mu}$ and $\nabla\bG(\theta^\star)^\top \bar{\Sigma}  \nabla\bG(\theta^\star))$.

For notation simplicity, we denote $\nabla^2 L = \nabla^2 L(\theta^\star, \blambda^\star), \nabla \bG = \nabla \bG(\theta^\star)$ and $H = (\nabla^2 L)^{-1} \nabla \bG [\nabla \bG^\top (\nabla^2 L)^{-1} \nabla \bG]^{-1}$. By block matrix inversion, we have
\[
B^{-1} = \begin{bmatrix}
    (\nabla^2 L)^{-1} - H \nabla \bG^\top (\nabla^2 L)^{-1} & H \\
    H^\top & -[\nabla \bG^\top (\nabla^2 L)^{-1} \nabla \bG]^{-1}.
\end{bmatrix}
\]
By \eqref{eq:solve-linear-system}, we have
\[
\bar{\mu} = -\left\{(\nabla^2 L)^{-1} - H \nabla \bG^\top (\nabla^2 L)^{-1}\right\} \mu_1 - H \mu_2.
\]
Note that $\nabla \bG^\top H = \bI_K$ and $\nabla \bG^\top \left\{(\nabla^2 L)^{-1} - H \nabla \bG^\top (\nabla^2 L)^{-1}\right\} = \zeros_{K\times K}$. We have 
\[
\nabla \bG(\theta^\star)^\top \bar{\mu} = -\nabla \bG^\top \left\{(\nabla^2 L)^{-1} - H \nabla \bG^\top (\nabla^2 L)^{-1}\right\} \mu_1 - \nabla \bG H \mu_2 = -\mu_2
\]

and
\begin{align*}
    &\nabla\bG(\theta^\star)^\top \bar{\Sigma}  \nabla\bG(\theta^\star) \\
    =& \nabla \bG^\top \left[\left\{(\nabla^2 L)^{-1} - H \nabla \bG^\top (\nabla^2 L)^{-1}\right\} \Sigma_{11} + H\Sigma_{21}\right]\underbrace{\left\{(\nabla^2 L)^{-1} - (\nabla^2 L)^{-1} \nabla \bG H^\top\right\} \nabla \bG}_{=\zeros_{K\times K}} \\
    &+ \underbrace{\nabla \bG^\top \left\{(\nabla^2 L)^{-1} - H \nabla \bG^\top (\nabla^2 L)^{-1}\right\}}_{=\zeros_{K\times K}} \Sigma_{12} H^\top \nabla \bG + \nabla \bG^\top H \Sigma_{22} H^\top \nabla \bG \\
    =& \Sigma_{22}.
\end{align*}

Therefore, we conclude that
\begin{equation*}
    \sqrt{n}
    \begin{bmatrix}
        \bbE_{P_0}[g_1(\htheta_n;Z)]\\
        \vdots \\
        \bbE_{P_0}[g_K(\htheta_n;Z)]
    \end{bmatrix} \dto \cN(-\mu_2, \Sigma_{22}) \overset{d}{=\joinrel=} \cN\left(-\begin{bmatrix}\sqrt{\rho_1\Var_{P_0}[g_1(\theta^\star;Z)]} \\ \vdots \\ \sqrt{\rho_K\Var_{P_0}[g_K(\theta^\star;Z)]}\end{bmatrix},\Var_{P_0}\begin{bmatrix}g_1(\theta^\star;Z)\\ \vdots \\ g_K(\theta^\star;Z)\end{bmatrix}\right).
\end{equation*}
Hence we complete the proof of Theorem \ref{thm:robust-SAA-multiple-expected-value-constraint-asymptotically-exact-constraint-satisfaction}. \hfill $\square$

\section{Proof of Corollary \ref{cor:robust-SAA-multiple-expected-value-constraint-asymptotically-exact-constraint-satisfaction}}
\label{sec:proof-cor-3.2}

Recall that $D = \diag(\Var_{P_0}[g_1(\theta^\star;Z)], \ldots, \Var_{P_0}[g_K(\theta^\star;Z)])$. According to Theorem \ref{thm:robust-SAA-multiple-expected-value-constraint-asymptotically-exact-constraint-satisfaction}, we have
\begin{align*}
    &\lim_{n\to\infty}\bbP\left\{\htheta_n~\text{is feasible}\right\} \\
    =& \lim_{n\to\infty}\bbP\left\{\begin{bmatrix}
        \bbE_{P_0}[g_1(\htheta_n;Z)]\\
        \vdots \\
        \bbE_{P_0}[g_K(\htheta_n;Z)]
    \end{bmatrix}  \in -\bbR_+^K \right\} \\
    =& \lim_{n\to\infty}\bbP\left\{\sqrt{n}\begin{bmatrix}
        \bbE_{P_0}[g_1(\htheta_n;Z)]\\
        \vdots \\
        \bbE_{P_0}[g_K(\htheta_n;Z)]
    \end{bmatrix}  \leq \zeros_K \right\} \\
    =& \bbP\left\{\cN\left(-\begin{bmatrix}\left(\rho_1\Var_{P_0}[g_1(\theta^\star;Z)]\right)^{\frac12} \\ \vdots \\ \left(\rho_K\Var_{P_0}[g_K(\theta^\star;Z)]\right)^{\frac12}\end{bmatrix},\Var_{P_0}\begin{bmatrix}g_1(\theta^\star;Z)\\ \vdots \\ g_K(\theta^\star;Z)\end{bmatrix}\right) \leq \zeros_K\right\} \\
    =& \bbP\left\{\cN\left(-D^{-\frac12}\begin{bmatrix}\left(\rho_1\Var_{P_0}[g_1(\theta^\star;Z)]\right)^{\frac12} \\ \vdots \\ \left(\rho_K\Var_{P_0}[g_K(\theta^\star;Z)]\right)^{\frac12}\end{bmatrix},D^{-\frac12}\Var\begin{bmatrix}g_1(\theta^\star;Z)\\ \vdots \\ g_K(\theta^\star;Z)\end{bmatrix}D^{-\frac12}\right) \leq D^{-\frac12}\zeros_K\right\} \\
    =& \bbP\left\{\cN\left(-\begin{bmatrix}\sqrt{\rho_1} \\ \vdots \\ \sqrt{\rho_K}\end{bmatrix},\operatorname{Corr}_{P_0}\begin{bmatrix}g_1(\theta^\star;Z)\\ \vdots \\ g_K(\theta^\star;Z)\end{bmatrix}\right) \leq \zeros_K\right\} \\
    =& \bbP\left\{\cN\left(\zeros_K,\operatorname{Corr}_{P_0}\begin{bmatrix}g_1(\theta^\star;Z)\\ \vdots \\ g_K(\theta^\star;Z)\end{bmatrix}\right) \leq \begin{bmatrix}\sqrt{\rho_1} \\ \vdots \\ \sqrt{\rho_K}\end{bmatrix}\right\}.
\end{align*}

Hence we complete the proof of Corollary \ref{cor:robust-SAA-multiple-expected-value-constraint-asymptotically-exact-constraint-satisfaction}. \hfill $\square$

\section{Dual ascent algorithm for \eqref{eq:robust-SAA-multiple-expected-value-constraint}}
\label{sec:dual-ascent-multiple-constraints}
We summarize the dual ascent algorithm for solving \eqref{eq:robust-SAA-multiple-expected-value-constraint} in Algorithm \ref{alg:robust-SAA-multiple-expected-value-constraint}. Similar to Algorithm \ref{alg:robust-SAA-single-expected-value-constraint}, the main cost of Algorithm \ref{alg:robust-SAA-multiple-expected-value-constraint} is incurred in the evaluation of dual function. The dual function evaluation step is still suitable for stochastic approximation. Therefore, the total cost of Algorithm \ref{alg:robust-SAA-multiple-expected-value-constraint} is comparable to that of a standard distributionally robust optimization problem.

\begin{algorithm*}
   \caption{Dual ascent algorithm for \eqref{eq:robust-SAA-multiple-expected-value-constraint}}
   \label{alg:robust-SAA-multiple-expected-value-constraint}
\begin{algorithmic}[1]
   \State {\bfseries Input:} starting dual iterate $\blambda_0 = (\lambda_{01},\ldots,\lambda_{0K})^\top\in\bbR^K_+$
   \Repeat
   \State Evaluate dual function: \[\textstyle\quad~~(\theta_t,\bmu_t,\bnu_t) \gets \argmin_{\theta,\bmu\in\bbR^K_+,\bnu}\frac1n\sum_{i=1}^n f(\theta;Z_i) + \textstyle\sum_{k=1}^K\lambda_{tk}\left\{\frac1n\sum_{i=1}^n\mu_{tk}\varphi^*(\frac{g_{k}(\theta;Z_i) - \nu_{tk}}{\mu_{tk}}) + \mu_{tk}\frac{\rho_k}{n} + \nu_{tk}\right\}\]
   \State Dual ascent update: \[\textstyle\lambda_{t+1,k} \gets \left[\lambda_{tk} + \eta_t\left\{\textstyle\frac1n\sum_{i=1}^n\mu_{tk}\varphi^*(\frac{g_{k}(\theta_t;Z_i) - \nu_{tk}}{\mu_{tk}}) + \mu_{tk}\frac{\rho_\alpha}{n} + \nu_{tk}\right\}\right]_+, k\in[K]\quad\quad\]
   \Until{converged}
\end{algorithmic}
\end{algorithm*}

\section{Simulations: details and more}
\label{sec:details-simulations}
In this section, we provide details for simulations on the multi-item newsvendor problem with independent constraints (which we present in the main text), and conduct more simulations on: (1) multi-item newsvendor problem with dependent constraints, and (2) single-item newsvendor problem with a single constraint.

\subsection{Multi-item newsvendor problem}
First recall that in Section \ref{sec:simulation}, we simulate the frequency of constraint satisfaction for the following multi-item newsvendor problem:
\begin{equation}
\label{eq:maximization-problem-multi-item-newsvendor-appendix}
\begin{array}{ll}
\max_{\theta\in\Theta} & \mathbb{E}_{P_0} \big[p^\top \min \{Z, \theta\}- c^\top\theta\big] \\
\subjectto & \bbE_{P_0}[(\|Z^{(1)}\|_2^2-\|\theta^{(1)}\|_2^2)_{+}] \leq \varepsilon_1 \\
 & \bbE_{P_0}[(\|Z^{(2)}\|_2^2-\|\theta^{(2)}\|_2^2)_{+}] \leq \varepsilon_2 \\
\end{array}
\end{equation}
where $c \in \bbR^d_{+}$ is the manufacturing cost, $p \in \bbR^d_{+}$ is the sell price, $\theta \in \Theta = [0,100]^d$ is the number of items in stock, $Z\in\bbR^d$ is a random variable with probability distribution $P_0$ representing the demand, and there are $d$ items in total. The distribution $P_0$ is unknown but we observe IID samples $Z_1,\ldots,Z_n$ from $P_0$. All of the items have been partitioned into two groups so that the corresponding demand and stock can be written as $Z = (Z^{(1)}, Z^{(2)})$ and $\theta = (\theta^{(1)}, \theta^{(2)})$. The constraints in the problem exclude stock levels that underestimate the demand too much for each group of items, where $\varepsilon_1,\varepsilon_2 > 0$ indicate tolerance level of such underestimation. The target of the problem is to maximize the profit while satisfying the constraints.

We can rewrite the maximization problem \eqref{eq:maximization-problem-multi-item-newsvendor-appendix} as a minimization problem with expected value constraints in the form of \eqref{eq:multiple-expected-value-constraint}, that is,
\begin{equation}
\label{eq:minimization-problem-multi-item-newsvendor-appendix}
\begin{array}{ll}
\min_{\theta\in\Theta} & \mathbb{E}_{P_0} \big[c^\top\theta - p^\top \min \{Z, \theta\}\big] \\
\subjectto & \bbE_{P_0}[(\|Z^{(1)}\|_2^2-\|\theta^{(1)}\|_2^2)_{+} - \varepsilon_1] \leq 0 \\
 & \bbE_{P_0}[(\|Z^{(2)}\|_2^2-\|\theta^{(2)}\|_2^2)_{+} - \varepsilon_2] \leq 0 \\
\end{array}
\end{equation}
so that we can apply our method \eqref{eq:robust-SAA-multiple-expected-value-constraint}.

We generate $Z_1,\ldots,Z_n$ IID from multivariate normal distribution $\cN(\mu, \Sigma)$. In addition, we set the number of items $d = 4$, the mean of the normal distribution $\mu = (10, 10, 10, 10)^\top$, the cost $c = (1, 1, 1, 1)^\top$, the price $p = (2, 2, 2, 2)^\top$, and the tolerance level of underestimation $(\varepsilon_1, \varepsilon_2) = (1, 1)$. Moreover, we partition the items into the group of the first two items and the group of the last two items. We solve the empirical problem with robust constraint:
\begin{equation}
\label{eq:robust-minimization-problem-multi-item-newsvendor-appendix}
\begin{array}{ll}
\min_{\theta\in\Theta} & \mathbb{E}_{\widehat{P}_n} \big[c^\top\theta - p^\top \min \{Z, \theta\}\big] \\
\subjectto & \sup_{D_\varphi (P\|\widehat{P}_n)\leq \rho/n} \bbE_{P}[(\|Z^{(1)}\|_2^2-\|\theta^{(1)}\|_2^2)_{+} - \varepsilon_1] \leq 0\\
& \sup_{D_\varphi (P\|\widehat{P}_n)\leq \rho/n} \bbE_{P}[(\|Z^{(2)}\|_2^2-\|\theta^{(2)}\|_2^2)_{+} - \varepsilon_2] \leq 0\\
\end{array}
\end{equation}
using samples of size $n \in \{30, 300, 3000\}$.

For the covariance of the multivariate normal distribution, we consider a exchangeable correlation structure
\begin{equation*}
    \Sigma = 9\times\begin{bmatrix}
        1 & r & r & r \\
        r & 1 & r & r \\
        r & r & 1 & r \\
        r & r & r & 1
    \end{bmatrix}
\end{equation*}

We solve \eqref{eq:robust-minimization-problem-multi-item-newsvendor-appendix} with $\brho = (\rho,\rho)^\top = (z_\alpha, z_\alpha)^\top$ for $\alpha \in \{0.4, 0.25, 0.1, 0.05, 0.005\}$, of which the corresponding $\sqrt{\rho}\in\{0.253, 0.674, 1.281, 1.644, 2.575\}$.

\paragraph{Independent constraints} In Section \ref{sec:simulation}, we consider $r = 0$ so that the two constraints are generally uncorrelated with each other. As suggested by the asymptotic theory in Section \ref{sec:multiple-expected-value-constraints}, the nominal probability of constraint satisfaction is $1-\alpha$ for each constraint and $(1-\alpha)^2$ for both constraints due to the independence of two constraints. The results are discussed in Section \ref{sec:simulation}.

\paragraph{Dependent constraints} Now we consider $r = 0.6$ so that the two constraints are generally correlated with each other. As suggested by the asymptotic theory in Section \ref{sec:multiple-expected-value-constraints}, the nominal probability of constraint satisfaction is $1-\alpha$ for each constraint. The nominal probability of constraint satisfaction for both constraints is no longer $(1-\alpha)^2$ due to the dependence of two constraints, but the probability is given by
\begin{equation*}
    \bbP\left\{\cN\left(\begin{bmatrix}
    0 \\
    0
    \end{bmatrix}, \begin{bmatrix}
    1 & \Corr_{P_0}(g_1(\theta^\star;Z), g_2(\theta^\star;Z)) \\
    \Corr_{P_0}(g_1(\theta^\star;Z), g_2(\theta^\star;Z)) & 1
    \end{bmatrix}\right) \leq_{\bbR^2} \begin{bmatrix}
    \sqrt{\rho} \\
    \sqrt{\rho}
    \end{bmatrix}\right\}.
\end{equation*}

\begin{figure}[h]
\centering
\begin{minipage}{0.32\textwidth}
  \centering
  \includegraphics[width=1\linewidth]{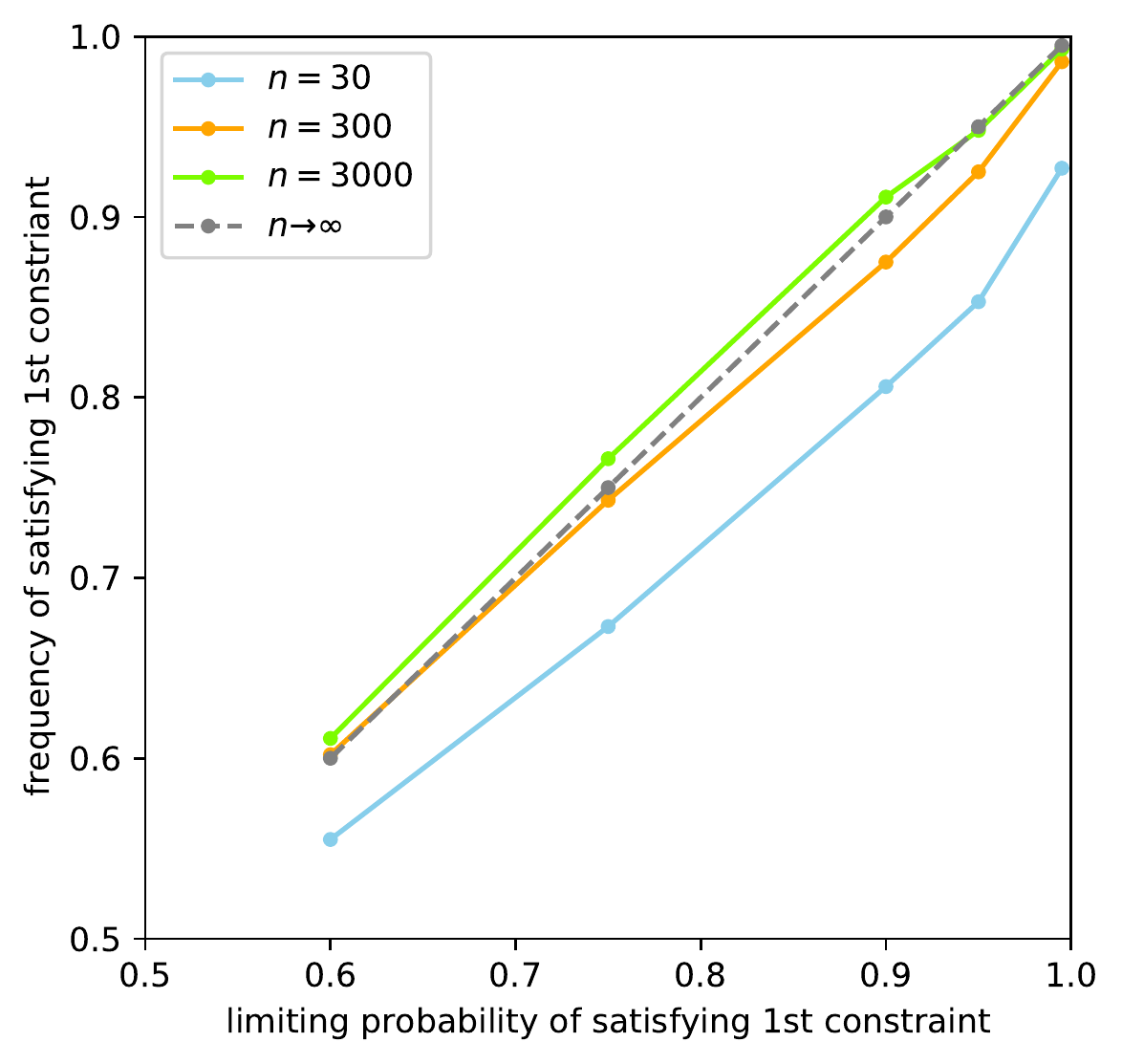}
\end{minipage}
\begin{minipage}{0.32\textwidth}
  \centering
  \includegraphics[width=1\linewidth]{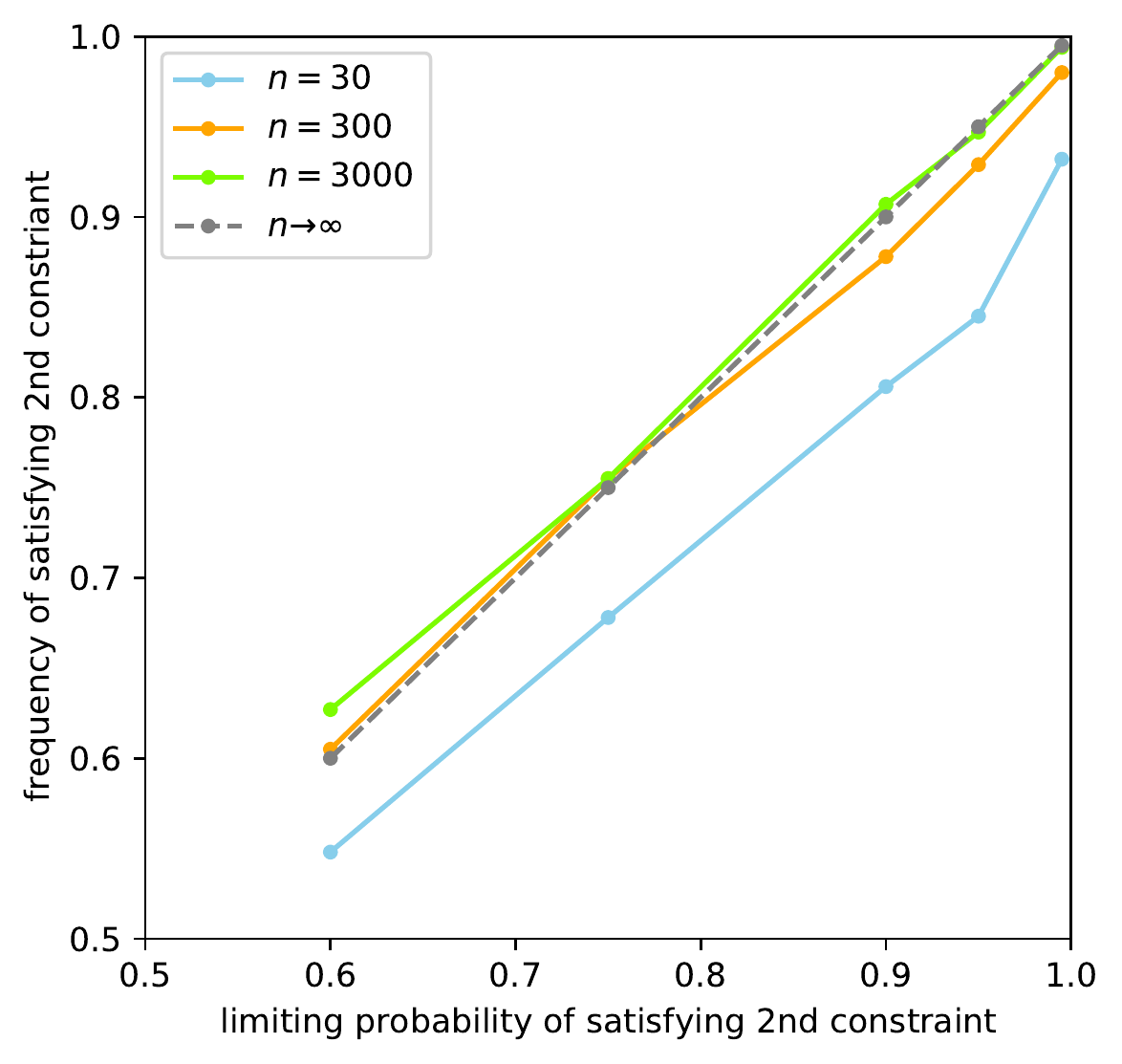}
\end{minipage}
\begin{minipage}{0.32\textwidth}
  \centering
  \includegraphics[width=1\linewidth]{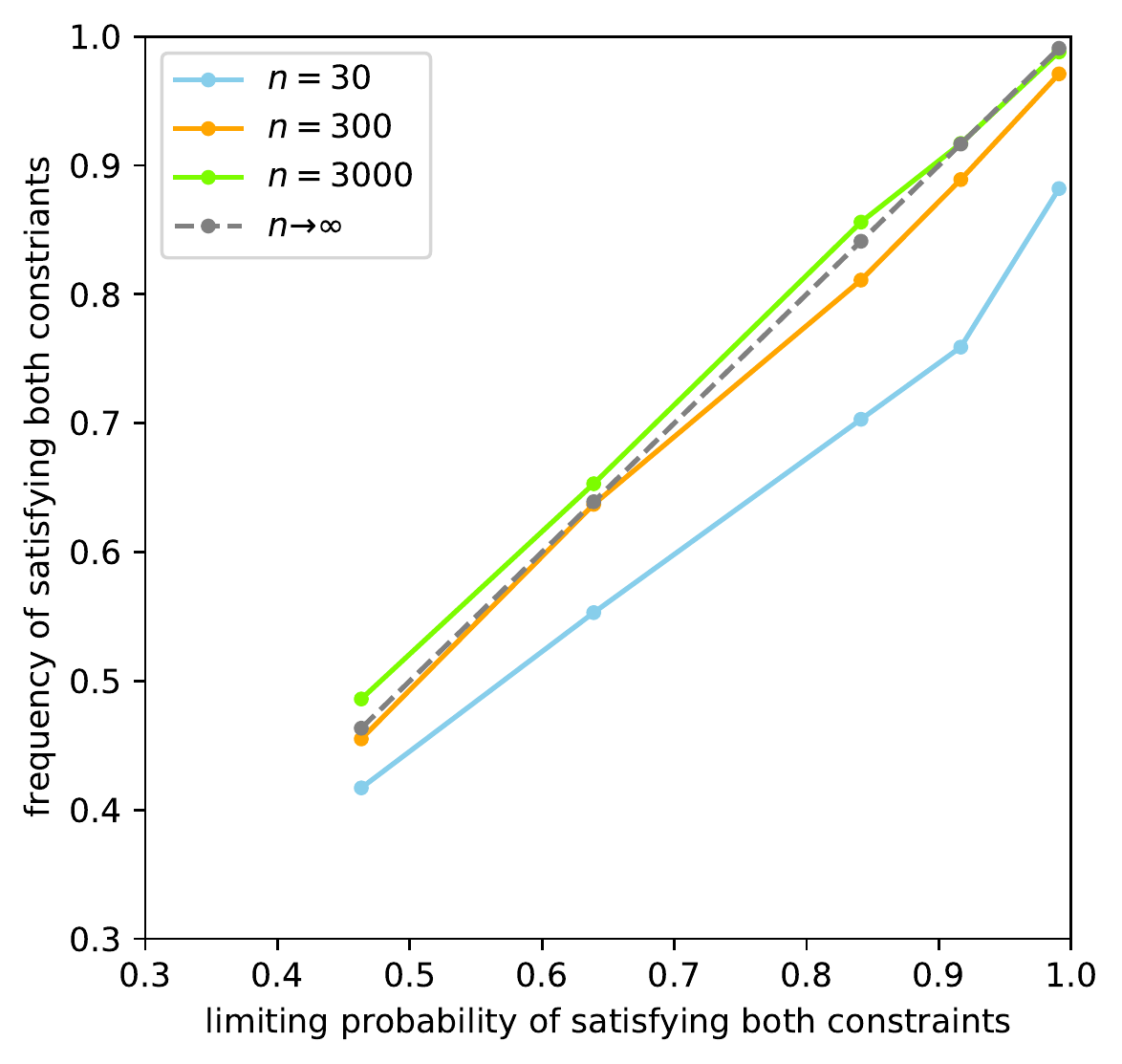}
\end{minipage}
\caption{Frequency versus limiting probability of constraint satisfaction of the first constraint (left), the second constraint (middle), and both of the constraints (right). }
\label{fig:3}
\end{figure}

In Figure \ref{fig:3}, we plot frequencies of constraint satisfaction for each constraint and both constraints, all of which are averaged over $1000$ replicates. As the sample size $n$ grows, the frequency versus probability curve converges to the theoretical dashed line of limiting probability of constraint satisfaction, validating our theory in the large sample regime. We note that in this example, the frequency of constraint satisfaction is higher than that of the experiments with independent constraints, for each $\brho$.

\subsection{Single-item newsvendor problem}
In this subsection, we consider the following single-item newsvendor problem:
\begin{equation*}
\begin{array}{ll}
\max_{\theta \in \Theta} & \mathbb{E}_{P_0}\big[p \min \{Z, \theta\}- c\theta\big] \\
\subjectto & \bbE_{P_0}\big[(Z-\theta)_{+}\big] \leq \varepsilon
\end{array}
\end{equation*}
where $c > 0$ is the manufacturing cost, $p \geq c$ is the sell price, $\theta \in \Theta = [0,100]$ is the number of items in stock, and $Z$ is a random variable with probability distribution $P_0$ representing the demand. The distribution $P_0$ is unknown but instead we observe IID samples $Z_1,\ldots,Z_n$ from $P$. The constraint in the problem excludes stocking levels that underestimate the demand too much, where $\varepsilon> 0$ indicates tolerance level of such underestimation. The target of the problem is to maximize the profit while satisfying the constraint. Note that the problem is equivalent to
\begin{equation*}
\begin{array}{ll}
\min_{\theta \in \Theta} & \mathbb{E}_{P_0}\big[c\theta - p \min \{Z, \theta\}\big] \\
\subjectto & \bbE_{P_0}\big[(Z-\theta)_{+} - \varepsilon\big] \leq 0
\end{array}
\end{equation*}
which is a particular case of \eqref{eq:single-expected-value-constraint}.

We generate $Z_1,\ldots,Z_n$ IID from exponential distribution with mean $10$. In addition, we set the cost $c = 1$, the price $p = 2$, and the tolerance level of underestimation $\varepsilon = 1$. We solve the empirical problem with robust constraint:
\begin{equation*}
\begin{array}{ll}
\min_{\theta \in \Theta} & \mathbb{E}_{\widehat{P}_n}\big[c\theta - p \min \{Z, \theta\}\big] \\
\subjectto & \sup_{D_\varphi (P\|\widehat{P}_n)\leq \rho/n} \bbE_{P}\big[(Z-\theta)_{+} - \varepsilon\big] \leq 0
\end{array}
\end{equation*}
using samples of size $n \in \{300, 3000, 30000\}$.

\begin{figure}[h]
    \centering
    \includegraphics[width=0.32\columnwidth]{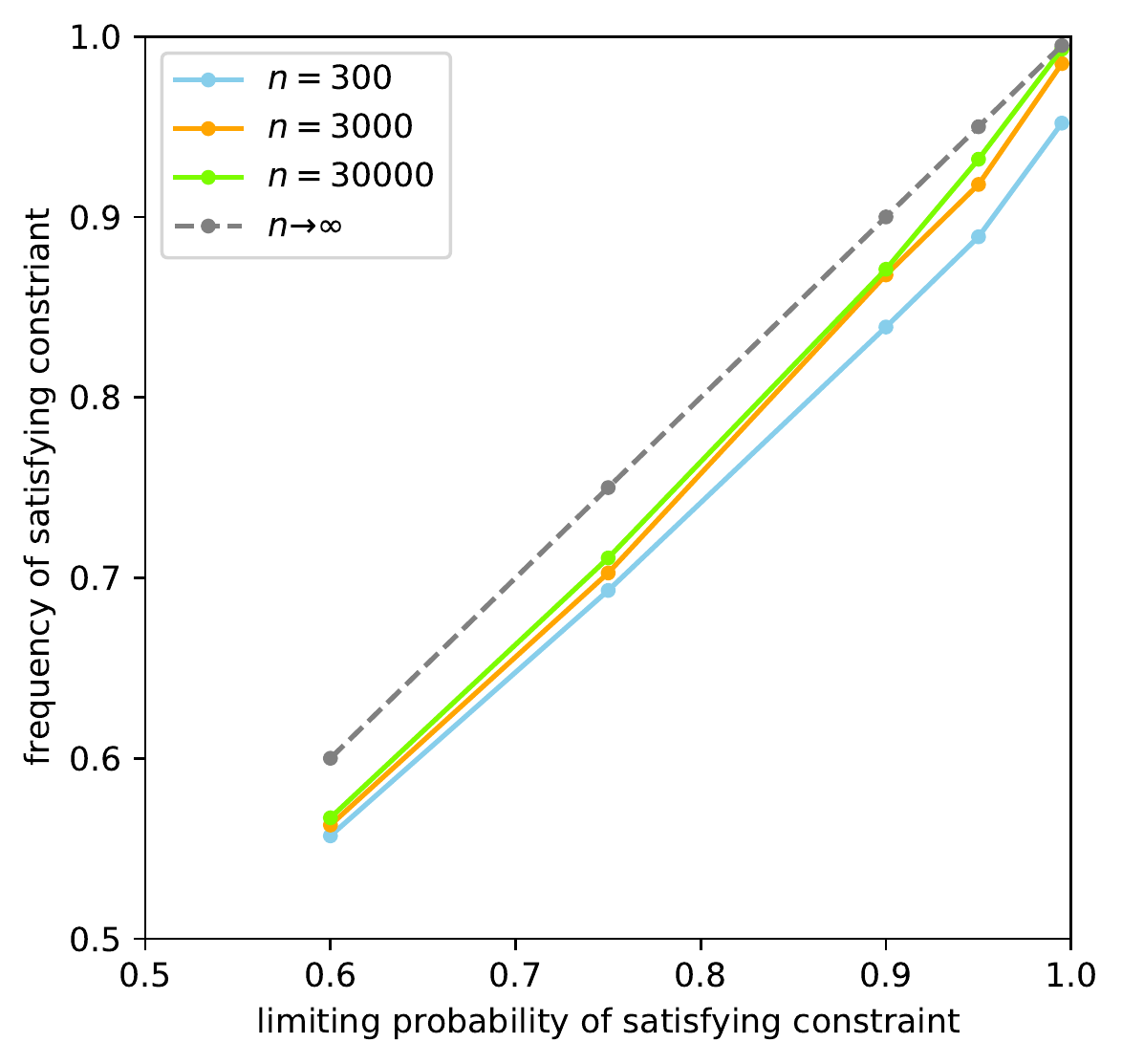}
    \caption{Frequency versus limiting probability of constraint satisfaction.}
    \label{fig:4}
\end{figure}

In Figure \ref{fig:4}, we plot frequencies of constraint satisfaction which are averaged over $1000$ replicates. As the sample size $n$ grows, the frequency versus probability curve converges to the theoretical dashed line of limiting probability of constraint satisfaction, validating our theory in the large sample regime.

\section{Application of our method to $\varepsilon$-equality of opportunity}
\label{sec:instantiate-EO}
Continuing with Section \ref{sec:unknown-active-set}, we demonstrate how to apply our method to enforce $\varepsilon$-equality of opportunity. To keep things simple, we assume there are only two demographic groups; \ie\ $|\cA| = 2$. Without loss of generality, we refer to one group as advantaged ($A=1$) and the other as disadvantaged ($A=0$). First we estimate rates $\bbP\left\{A = 1, Y = 1\right\}$ and $\bbP\left\{A = 0, Y = 1\right\}$ consistently by
\begin{equation*}
    \textstyle\widehat{p}_1 = \frac{1}{n}\sum_{i=1}^n \ones\left\{A_i = 1, Y_i = 1\right\}\quad\text{and}\quad\widehat{p}_0 = \frac{1}{n}\sum_{i=1}^n \ones\left\{A_i = 0, Y_i = 1\right\}.
\end{equation*}
Then, we construct a robust constraint
\begin{equation*}
    \textstyle\sup_{P:D_\varphi (P\|\hP_n)\leq \rho/n} \bbE_{(X,A,Y)\sim P}\left[\frac{\ones\{\widehat{Y} = 1, A = 1, Y = 1\}}{\widehat{p}_1} - \frac{\ones\{\widehat{Y} = 1, A = 0, Y = 1\}}{\widehat{p}_0} -\varepsilon\right] \leq 0,
\end{equation*}
or equivalently
\begin{equation*}
    \textstyle\sup_{\bp:\sum_{i=1}^n \varphi(np_i)\leq \rho} \sum_{i=1}^n p_i\left[\frac{\ones\{f_\theta(X_i) = 1, A_i = 1, Y_i = 1\}}{\widehat{p}_1} - \frac{\ones\{f_\theta(X_i) = 1, A_i = 0, Y_i = 1\}}{\widehat{p}_0} -\varepsilon\right] \leq 0.
\end{equation*}

\section{Two-stage method for unknown active set}
\label{sec:details-unknown-active-set}

In this section, we show that the two-stage method in Section \ref{sec:unknown-active-set} also has the calibration property (similar to Theorem \ref{thm:robust-SAA-multiple-expected-value-constraint-asymptotically-exact-constraint-satisfaction} and Corollary \ref{cor:robust-SAA-multiple-expected-value-constraint-asymptotically-exact-constraint-satisfaction}) if the true program $(\cP_0)$, \ie\ \eqref{eq:multiple-expected-value-constraint}, is not ill-behaved.

First we recall the two-stage method:
\begin{enumerate}
    \item At the first stage, we solve the program $(\cP_n)$ with $\brho = \zeros_K$, that is, the sample average approximation (SAA) problem. We identify that $\widehat{J}_{+,n}\subset [K]$ is the active set of the SAA problem, that is, the $j$-th constraint of the SAA problem is active if and only if $j\in\widehat{J}_{+,n}$.
    \item At the second stage, we solve the program $(\cP_n)$ with $\brho$ such that $\rho_j$ is some positive number if $j \in \widehat{J}_{+,n}$. This means that at the second stage we replace the sample mean approximation to the constraint by its distributionally robust counterpart if such constraint was identified as active at the first stage.
\end{enumerate}

The index set $[K]$ can be partitioned into three parts:
\begin{align*}
    J_{+}(\theta^\star, \blambda^\star) &= \{k\in[K]: 
    \bbE_{P_0}[g_k(\theta^\star; Z)] = 0, \lambda^\star_k > 0\}, \\
    J_{0}(\theta^\star, \blambda^\star) &= \{k\in[K]: 
    \bbE_{P_0}[g_k(\theta^\star; Z)] = 0, \lambda^\star_k = 0\}, \\
    J_{-}(\theta^\star, \blambda^\star) &= \{k\in[K]: 
    \bbE_{P_0}[g_k(\theta^\star; Z)] < 0, \lambda_k^\star = 0\},
\end{align*}
where $J_+$ is the active set with positive Lagrange multipliers, $J_0$ is the active set with zero Lagrange multipliers, and $J_-$ is the inactive set. We assume the \emph{strict complementarity} holds in the sense that $J_0(\theta^\star, \blambda^\star) = \varnothing$.

\begin{proposition}[Preservance of active constraints]
\label{prop:preservance-of-active-constraints}
Assume the strict complementarity holds. We have $\widehat{J}_{+,n} = J_+(\theta^\star, \blambda^\star)$ with probability converging to $1$ as $n\to\infty$.
\end{proposition}

\emph{Proof of Proposition \ref{prop:preservance-of-active-constraints}.} Let $\htheta_n^{(\operatorname{SAA})}$ be a solution to the sample average approximation problem at the first stage. Let $\widehat{\blambda}_n$ be the associated Lagrange multiplier. It is known by the consistency of SAA \cite{rubinstein1993discrete} that
\begin{equation*}
    \widehat{\lambda}_{k,n} = \lambda^\star_k + o_p(1)~\text{for}~k\in J_+(\theta^\star, \blambda^\star)
\end{equation*}
and
\begin{equation*}
    \bbE_P[g_k(\htheta_n^{(\operatorname{SAA})}; Z)] = \bbE_P[g_k(\theta^\star; Z)] + o_p(1)~\text{for}~k\in J_{-}(\theta^\star, \blambda^\star)
\end{equation*}
Therefore, with probability converging to $1$, $\widehat{\lambda}_{k,n} > 0$, $k \in J_+(\theta^\star, \blambda^\star)$ and $\bbE_P[g_k(\htheta_n^{(\operatorname{SAA})}; Z)] < 0$, $k \in J_{-}(\theta^\star, \blambda^\star)$. Hence we complete the proof of Proposition \ref{prop:preservance-of-active-constraints}. \hfill $\square$

\begin{theorem}
\label{thm:unknown-active-set}
Suppose the true program $(\cP_0)$ has $m$ active constraints with positive Lagrange multipliers and $K-m$ inactive constraints (without loss of generality we let $J_+ = \{1,\ldots, m\}$ and $J_{-} = \{m+1, \ldots, K\}$). Let $\htheta_n$ be the two-stage estimator. Under the standing assumptions, we have
\begin{equation*}
    \lim_{n\to\infty}\bbP\left\{\begin{bmatrix}
        \bbE_{P_0}[g_1(\htheta_n;Z)]\\
        \vdots \\
        \bbE_{P_0}[g_K(\htheta_n;Z)]
    \end{bmatrix}  \in -\bbR_+^K \right\} = \bbP\left\{\cN\left(\zeros_m,\operatorname{Corr}_{P_0}\begin{bmatrix}g_1(\theta^\star;Z)\\ \vdots \\ g_m(\theta^\star;Z)\end{bmatrix}\right) \leq \begin{bmatrix}\sqrt{\rho_1} \\ \vdots \\ \sqrt{\rho_m}\end{bmatrix}\right\}.
\end{equation*}
\end{theorem}

This result shows that the limiting probability of satisfying the true constraints only depends on the correlation structure between active constraints and the uncertainty set radii for the active constraints.

\emph{Proof of Theorem \ref{thm:unknown-active-set}.} At the first stage, we identify $J \subset[K]$ as active set with probability $p_J$. Here the randomness is introduced by the data samples $Z_1,\ldots,Z_n$. Let $2^{[K]}$ be the power set of $[K]$. We have
\[
\sum_{J\in2^{[K]}} p_J = 1.
\]
By Proposition \ref{prop:preservance-of-active-constraints}, as $n\to\infty$ we have
\[
p_{J_+} \to 1
\]
and
\[
p_J \to 0~\text{for any}~J\in 2^{[K]}~\text{and}~J \neq J_+.
\]

By Corollary \ref{cor:robust-SAA-multiple-expected-value-constraint-asymptotically-exact-constraint-satisfaction}, we have
\begin{equation*}
    \lim_{n\to\infty}\bbP\left\{\htheta_n~\text{is feasible}\mid \text{identify}~J_+~\text{as active set}\right\} = \bbP\left\{\cN\left(\zeros_m,\operatorname{Corr}_{P_0}\begin{bmatrix}g_1(\theta^\star;Z)\\ \vdots \\ g_m(\theta^\star;Z)\end{bmatrix}\right) \leq \begin{bmatrix}\sqrt{\rho_1} \\ \vdots \\ \sqrt{\rho_m}\end{bmatrix}\right\}.
\end{equation*}
Therefore,
\begin{align*}
    &\bbP\left\{\htheta_n~\text{is feasible}\right\} \\
    =& \sum_{J\in 2^{[K]}} \bbP\left\{\htheta_n~\text{is feasible}\mid \text{identify}~J~\text{as active set}\right\} \bbP\left\{\text{identify}~J~\text{as active set}\right\} \\
    =& \underbrace{p_{J_+}}_{\to 1} \bbP\left\{\htheta_n~\text{is feasible}\mid \text{identify}~J_+~\text{as active set}\right\} + \sum_{J\neq J_+} \underbrace{p_J \bbP\left\{\htheta_n~\text{is feasible}\mid \text{identify}~J~\text{as active set}\right\}}_{\to 0} \\
    \to& \bbP\left\{\cN\left(\zeros_m,\operatorname{Corr}_{P_0}\begin{bmatrix}g_1(\theta^\star;Z)\\ \vdots \\ g_m(\theta^\star;Z)\end{bmatrix}\right) \leq \begin{bmatrix}\sqrt{\rho_1} \\ \vdots \\ \sqrt{\rho_m}\end{bmatrix}\right\}.
\end{align*}
Hence we complete the proof of Theorem \ref{thm:unknown-active-set}. \hfill $\square$

\section{Proxy dual ascent algorithm for non-differentiable constraints}
\label{sec:proxy-dual-ascent-algorithm}
We summarize in Algorithm \ref{alg:proxy-dual-ascent-single-constraints} the proxy dual ascent algorithm for solving a stochastic optimization problem with single non-differentiable constraint. The difference between Algorithm \ref{alg:proxy-dual-ascent-single-constraints} and \ref{alg:robust-SAA-single-expected-value-constraint} is in the step of evaluating (proxy) dual function: Algorithm \ref{alg:proxy-dual-ascent-single-constraints} uses a differentiable surrogate $\Tilde{g}$ (highlighted in \color{orange}orange\color{black}) instead of the non-differentiable $g$ in Algorithm \ref{alg:robust-SAA-single-expected-value-constraint} in this step.

\begin{algorithm*}
   \caption{Proxy dual ascent algorithm for single non-differentiable constraint}
   \label{alg:proxy-dual-ascent-single-constraints}
\begin{algorithmic}[1]
   \State {\bfseries Input:} starting dual iterate $\lambda_0 \ge 0$
   \Repeat
   \State Evaluate proxy dual function: \[\textstyle\quad~~(\theta_t,\mu_t,\nu_t) \gets \argmin_{\theta,\mu\ge 0,\nu}\frac1n\sum_{i=1}^n f(\theta;Z_i) + \lambda_t\left\{\textstyle\frac1n\sum_{i=1}^n\mu\varphi^*\big(\frac{\color{orange}\Tilde{g}\color{black}(\theta;Z_i) - \nu}{\mu}\big) + \mu\frac{\rho_\alpha}{n} + \nu\right\}\]
   \State Dual ascent update: $\lambda_{t+1} \gets \left[\lambda_t + \eta_t\left\{\textstyle\frac1n\sum_{i=1}^n\mu_t\varphi^*(\frac{g(\theta_t;Z_i) - \nu_t}{\mu_t}) + \mu_t\frac{\rho_\alpha}{n} + \nu_t\right\}\right]_+$
   \Until{converged}
\end{algorithmic}
\end{algorithm*}

Algorithm \ref{alg:proxy-dual-ascent-multiple-constraints} summarizes the proxy dual ascent algorithm for solving a stochastic optimization problem with multiple non-differentiable constraints. In contrast to Algorithm \ref{alg:robust-SAA-multiple-expected-value-constraint}, Algorithm \ref{alg:proxy-dual-ascent-multiple-constraints} replaces the non-differentiable functions $g_k$'s by their differentiable surrogates $\Tilde{g}_k$'s (highlighted in \color{orange}orange\color{black}) in the (proxy) dual function evaluation step.

\begin{algorithm*}
   \caption{Proxy dual ascent algorithm for multiple non-differentiable constraints}
   \label{alg:proxy-dual-ascent-multiple-constraints}
\begin{algorithmic}[1]
   \State {\bfseries Input:} starting dual iterate $\blambda_0 = (\lambda_{01},\ldots,\lambda_{0K})^\top\in\bbR^K_+$
   \Repeat
   \State Evaluate proxy dual function: \[\textstyle\quad~~(\theta_t,\bmu_t,\bnu_t) \gets \argmin_{\theta,\bmu\in\bbR^K_+,\bnu}\frac1n\sum_{i=1}^n f(\theta;Z_i) + \textstyle\sum_{k=1}^K\lambda_{tk}\left\{\frac1n\sum_{i=1}^n\mu_{tk}\varphi^*(\frac{\color{orange}\Tilde{g}_{k}\color{black}(\theta;Z_i) - \nu_{tk}}{\mu_{tk}}) + \mu_{tk}\frac{\rho_k}{n} + \nu_{tk}\right\}\]
   \State Dual ascent update: \[\textstyle\lambda_{t+1,k} \gets \left[\lambda_{tk} + \eta_t\left\{\textstyle\frac1n\sum_{i=1}^n\mu_{tk}\varphi^*(\frac{g_{k}(\theta_t;Z_i) - \nu_{tk}}{\mu_{tk}}) + \mu_{tk}\frac{\rho_\alpha}{n} + \nu_{tk}\right\}\right]_+, k\in[K]\quad\quad\]
   \Until{converged}
\end{algorithmic}
\end{algorithm*}

\section{Real data experiments: details and more}
\label{sec:details-experiments}

In this section, we provide details for experiments on Adult dataset with $\varepsilon$-demographic parity as the fairness goal (which we present in the main text), and conduct additional experiments on Adult dataset with $\varepsilon$-false positive rate parity as the fairness goal.

\subsection{Adult with $\varepsilon$-demographic parity}
\label{sec:adult-DP}

Recall that in Section \ref{sec:adult} the fairness goal is $\varepsilon$-demographic parity ($\varepsilon$-DP):
\[
\big|\bbP(\hY = 1\mid A = 1) - \bbP(\hY = 1\mid A = 0)\big|\leq \varepsilon,
\]
where $A = 1$ for male is the advantaged group and $A = 0$ for female is the disadvantaged group. We use a logistic regression model for classification by predicting $\widehat{Y} = \ones\{\theta^\top X \geq 0\}$ and training such model parameterized in $\theta$ by logistic loss. We implement the two-stage method and proxy dual ascent algorithm by replacing indicator function by the sigmoidal function $h_1(t) = (1 + e^{-at})^{-1}$ for $a = 2$. We do bootstrap evaluation by treating the full dataset as the true probability measure and sample such probability measure with replacement as the training distribution (with sampling rate $50\%$).

\subsection{Adult with $\varepsilon$-false positive rate parity}
\label{sec:adult-FPR}

Similar to the preceding subsection, now we consider the fairness goal to be $\varepsilon$-false positive rate parity:
\[
\big|\bbP(\hY = 1\mid Y = 0, A = 1) - \bbP(\hY = 1\mid  Y = 0, A = 0)\big|\leq \varepsilon,
\]
where $A = 1$ for male is the advantaged group and $A = 0$ for female is the disadvantaged group.

\begin{figure}[h]
\centering
\begin{minipage}{0.48\textwidth}
  \centering
  \includegraphics[width=1\linewidth]{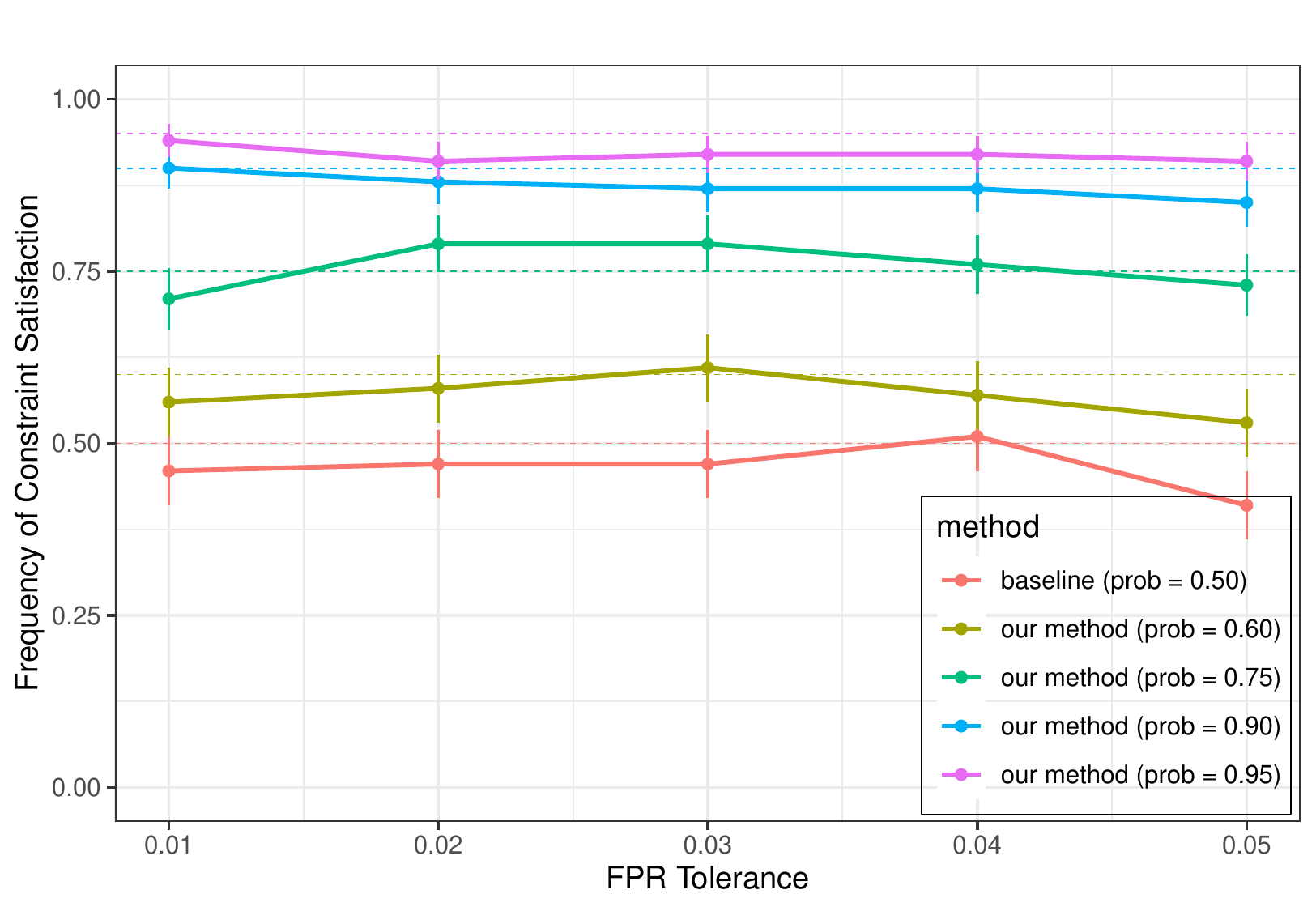}
\end{minipage}
\begin{minipage}{0.48\textwidth}
  \centering
  \includegraphics[width=1\linewidth]{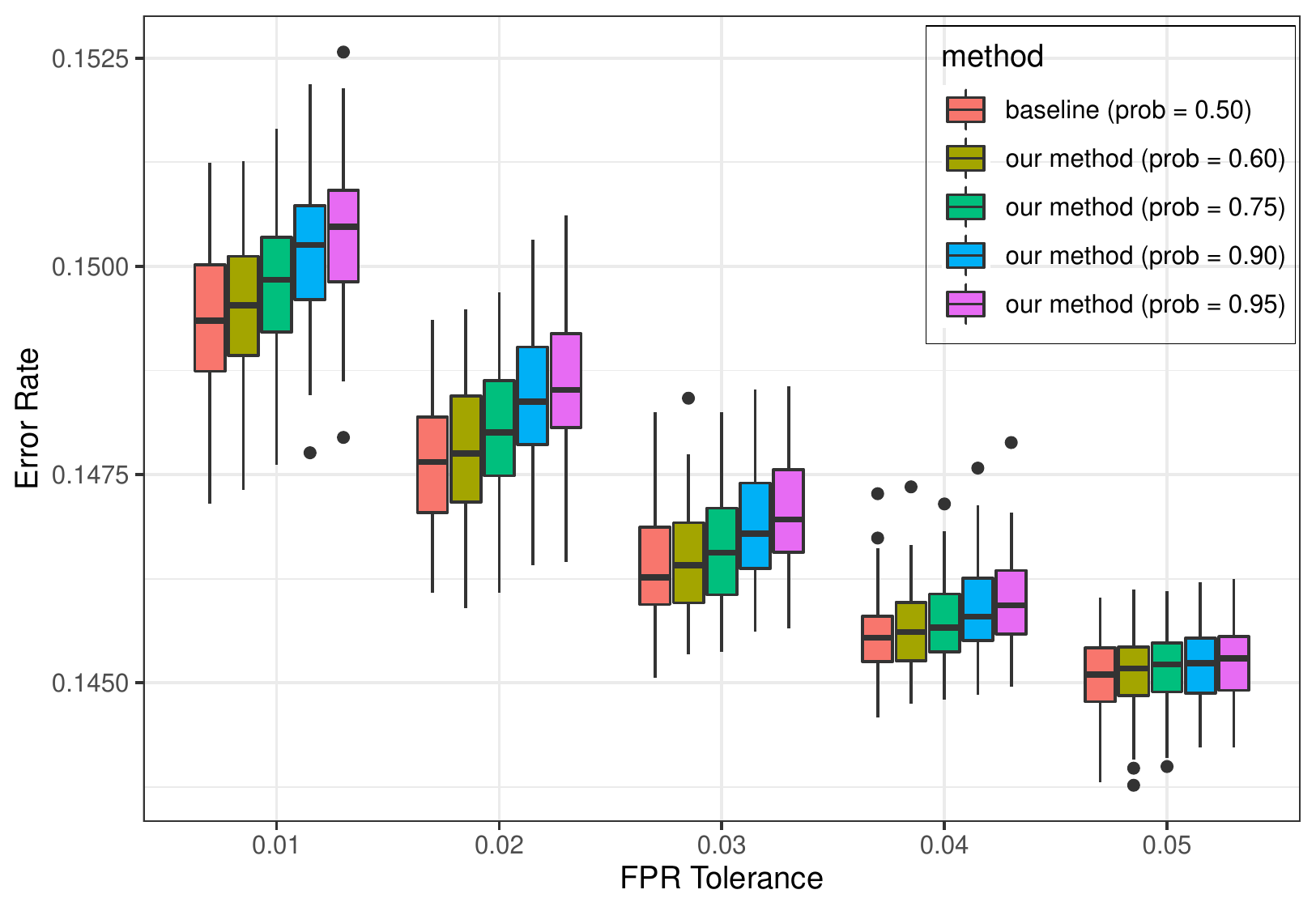}
\end{minipage}
\caption{Frequency of constraint satisfaction (left) and classification error rate (right) for different false positive rate parity tolerance $\varepsilon \in \{0.01, 0.02, 0.03, 0.04, 0.05\}$. Baseline (sample average approximation, SAA) and our methods (with nominal probability $0.60, 0.75, 0.90,0.95$) are compared.}
\label{fig:5}
\end{figure}

In Figure \ref{fig:5}, we have line plots for frequency of constraint satisfaction and box plots for classification error rate, all of which are summarized over $100$ replicates. The patterns shown in the left and right panels are similar to that in Figure \ref{fig:2}. We observe more variation in frequencies of constraint satisfaction across different false positive rate tolerance designs due to the label and demographic attribute imbalance of Adult dataset.

\section{Dual form of the robust constraint function \eqref{eq:robust-expected-value-constraint-dual-form}}
\label{sec:standard-dro-dual}
In this section, we provide a standard derivation for the dual form of the robust constraint function \eqref{eq:robust-expected-value-constraint-dual-form}.

We introduce the likelihood ratio $L(Z) = dP(Z)/d\hP_n(Z)$. By change of variable, we can rewrite the robust constraint function \eqref{eq:robust-expected-value-constraint-dual-form} as
\[
\sup\nolimits_{P:D_\varphi(P\| \hP_n)\leq\rho}\bbE_P\big[g(\theta;Z)\big] \textstyle=\sup_{L\geq 0}\{\bbE_{\hP_n}\big[L(Z)g(\theta;Z)\big]\mid\bbE_{\hP_n}\big[\varphi(L(Z))\big]\leq\rho, \bbE_{\hP_n}\big[L(Z)\big] = 1\},
\]
where the supremum takes over measurable functions. This gives us a constrained optimization problem. Let $\mu\geq 0$ be the Lagrange multiplier for $\bbE_{\hP_n}\big[\varphi(L(Z))\big]\leq\rho$ and $\nu\in\bbR$ be the Lagrange multiplier for $\bbE_{\hP_n}\big[L(Z)\big] = 1$. The corresponding Lagrangian is 
\begin{equation*}
    \cL(L,\mu,\nu) = \bbE_{\hP_n}\big[(g(\theta;Z) -\nu)L(Z) - \mu\varphi(L(Z))\big] + \mu\rho + \nu.
\end{equation*}

For regular $\varphi$-divergence, we have
\begin{align*}
    &\sup\nolimits_{P:D_\varphi(P\| \hP_n)\leq\rho}\bbE_P\big[g(\theta;Z)\big] \\
    =& \inf_{\mu\geq 0, \nu\in\bbR} \sup_{L\geq 0} \cL(L,\mu,\nu) \\
    =& \inf_{\mu\geq 0, \nu\in\bbR} \sup_{L\geq 0} \left\{\sum_{i=1}^n\big[(g(\theta;Z_i) - \nu) L(Z_i) - \mu \varphi(L(Z_i))\big] + \mu\rho + \nu\right\} \\
    =& \inf_{\mu\geq 0, \nu\in\bbR} \sup_{L\geq 0} \left\{\sum_{i=1}^n\mu\left[\frac{g(\theta;Z_i) - \nu}{\mu} L(Z_i) - \varphi(L(Z_i))\right] + \mu\rho + \nu\right\} \\
    =& \inf_{\mu\geq 0, \nu\in\bbR} \left\{\sum_{i=1}^n \mu \sup_{t_i\geq 0}\left\{\frac{g(\theta;Z_i) - \nu}{\mu} t_i - \varphi(t_i)\right\} + \mu\rho + \nu\right\} \\
    &= \inf_{\mu\geq 0, \nu\in\bbR} \left\{\sum_{i=1}^n \mu \varphi^*\left(\frac{g(\theta;Z_i) - \nu}{\mu}\right) + \mu\rho + \nu\right\}.
\end{align*}
Here the last equality holds according to the definition of the convex conjugate $\varphi^*(\cdot)$.

\section{Experiments on additional baseline and dataset}
\label{sec:more-baseline}
In this section, we conduct more experiments using two-dataset approach of \citep{cotter2019training} as an additional baseline and default of credit card clients dataset from UCI \citep{Dua:2019} as an additional dataset.

\subsection{Adult with $\varepsilon$-demographic parity (continued)}
We continue the experiments on Adult dataset with $\varepsilon$-demographic parity in Section \ref{sec:adult} and \ref{sec:adult-DP} by adding two-dataset approach of \citep{cotter2019training} as an additional baseline.

\begin{figure}[h]
\centering
\begin{minipage}{0.48\textwidth}
  \centering
  \includegraphics[width=1\linewidth]{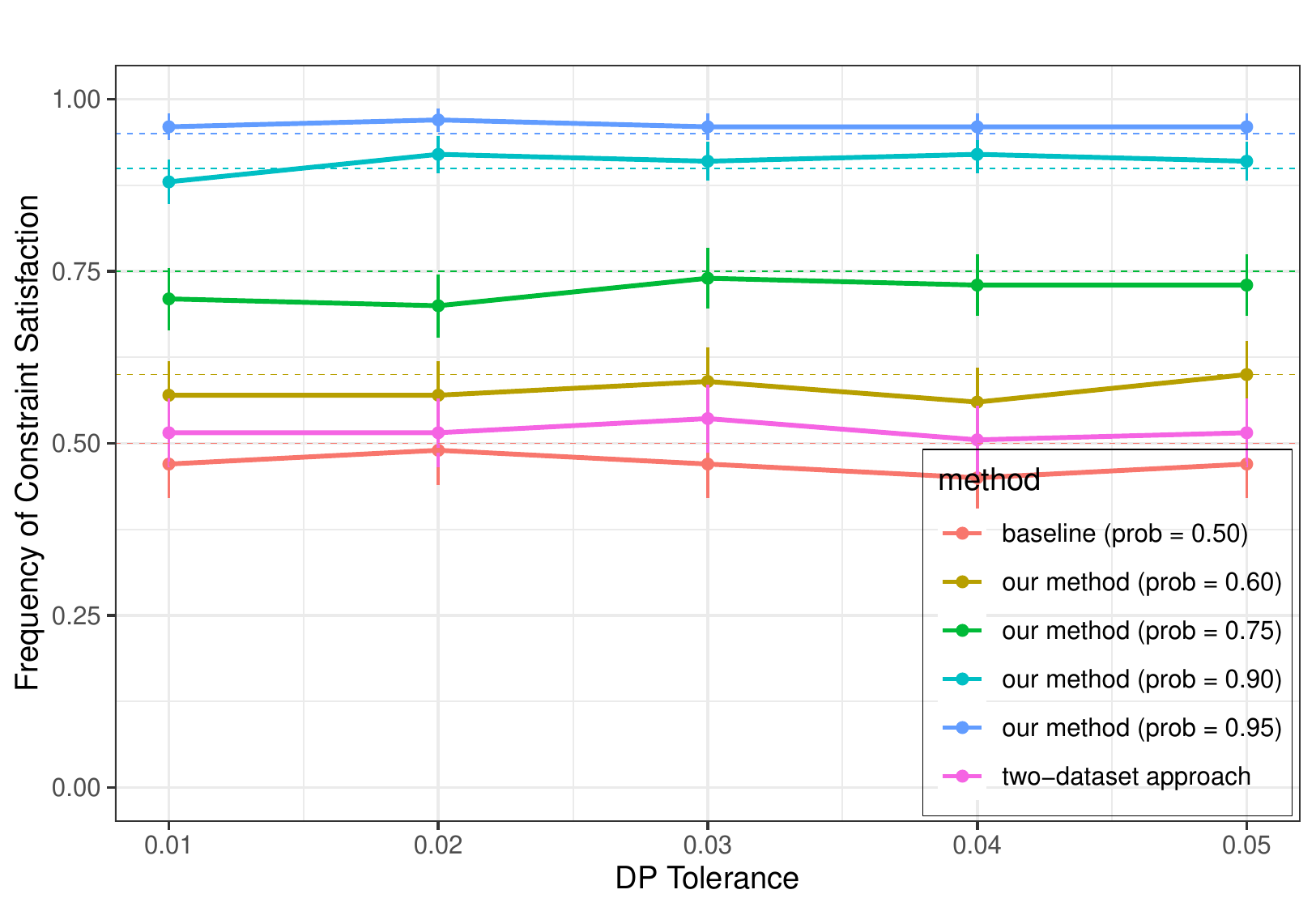}
\end{minipage}
\begin{minipage}{0.48\textwidth}
  \centering
  \includegraphics[width=1\linewidth]{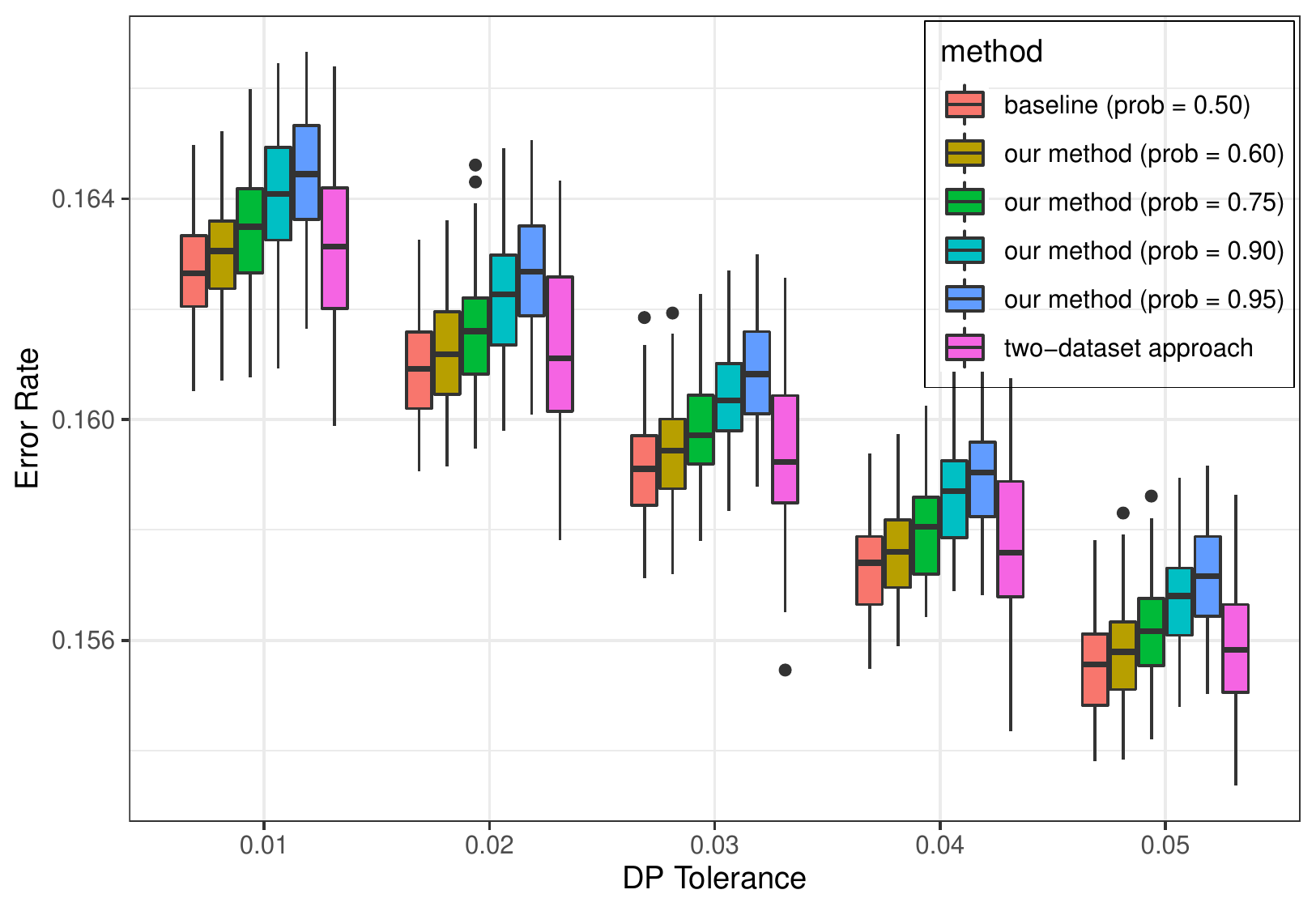}
\end{minipage}
\caption{Frequency of constraint satisfaction (left) and classification error rate (right) for different demographic parity tolerance $\varepsilon \in \{0.01, 0.02, 0.03, 0.04, 0.05\}$. Baseline (sample average approximation, SAA), our methods (with nominal probability $0.60, 0.75, 0.90,0.95$), and two-dataset approach \citep{cotter2019training} are compared.}
\label{fig:6}
\end{figure}

The two dataset approach of \citep{cotter2019training} splits the training set into two parts, one for updating the model parameters and the other for updating the Lagrangian multipliers, with the goal to improve the generalization of fairness constraints.
Figure \ref{fig:6} demonstrates that the two-dataset approach marginally improves the frequency of constraint satisfaction over the SAA.
The cost of such an improvement is increased variation in classification error rate. 
Due to the fact that the two-dataset approach does not use the entire training set to update the model parameters, statistical efficiency is sacrificed. 
Although the two-dataset approach outperforms the SAA in terms of constraint satisfaction frequency, both are inferior to our methods, which achieve the user-prescribed level of frequency.

\subsection{Adult with $\varepsilon$-false positive rate parity (continued)}
We continue the experiments on Adult dataset with $\varepsilon$-false positive rate parity in Section \ref{sec:adult-FPR} by adding two-dataset approach of \citep{cotter2019training} as an additional baseline.

\begin{figure}[h]
\centering
\begin{minipage}{0.48\textwidth}
  \centering
  \includegraphics[width=1\linewidth]{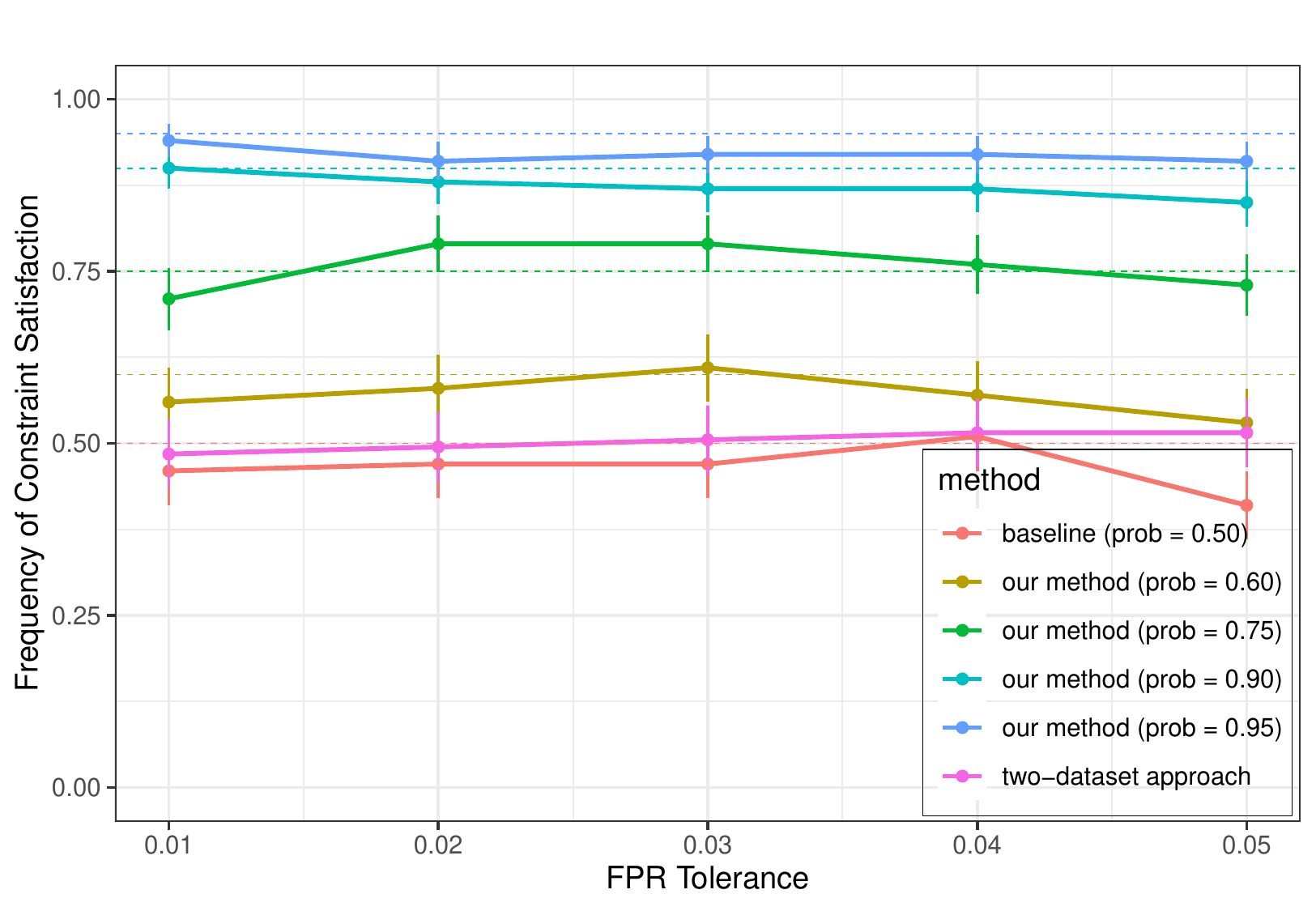}
\end{minipage}
\begin{minipage}{0.48\textwidth}
  \centering
  \includegraphics[width=1\linewidth]{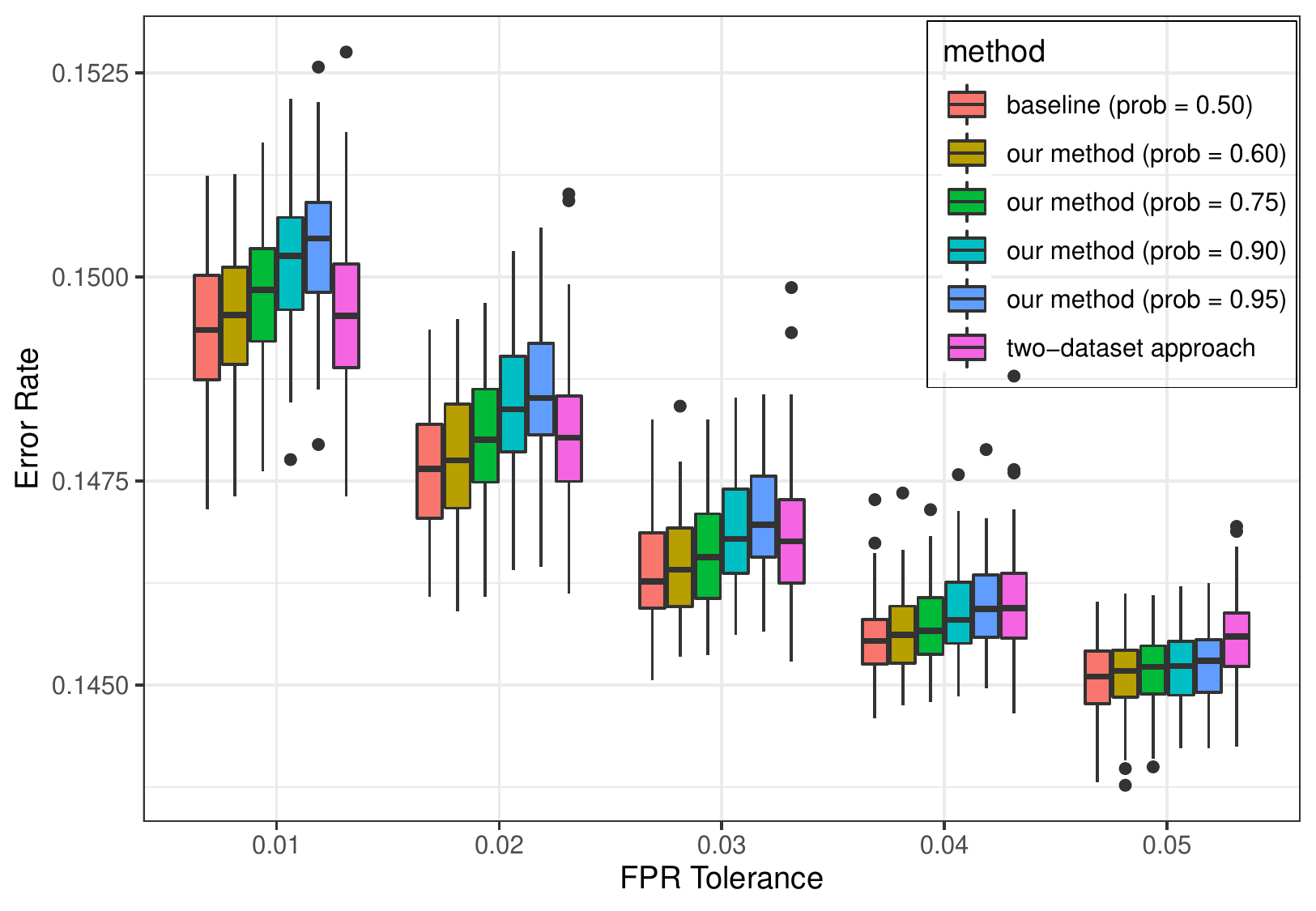}
\end{minipage}
\caption{Frequency of constraint satisfaction (left) and classification error rate (right) for different false positive rate parity tolerance $\varepsilon \in \{0.01, 0.02, 0.03, 0.04, 0.05\}$. Baseline (sample average approximation, SAA), our methods (with nominal probability $0.60, 0.75, 0.90,0.95$), and two-dataset approach \citep{cotter2019training} are compared.}
\label{fig:7}
\end{figure}

Figure \ref{fig:7} demonstrates similar patterns as Figure \ref{fig:6}. The two-dataset approach improves constraint satisfaction frequency over the SAA, but has a worse classification error rate or one comparable to it. Our methods achieve desired frequency of constraint satisfaction at the user-prescribed level.

\subsection{Credit with $\varepsilon$-demographic parity}
Predicting whether or not an individual defaulted on a loan is the classification task of the UCI default of credit card clients (Credit) dataset. Membership in the demographic group is determined by an individual's level of education: $A = 1$ if a person has earned a graduate degree; otherwise, $A = 0$.

We consider the fairness goal to be $\varepsilon$-demographic parity:
\[
\big|\bbP(\hY = 1\mid A = 1) - \bbP(\hY = 1\mid A = 0)\big|\leq \varepsilon,
\]
where $A = 1$ for individuals with a graduate degree is the advantaged group.

\begin{figure}[h]
\centering
\begin{minipage}{0.48\textwidth}
  \centering
  \includegraphics[width=1\linewidth]{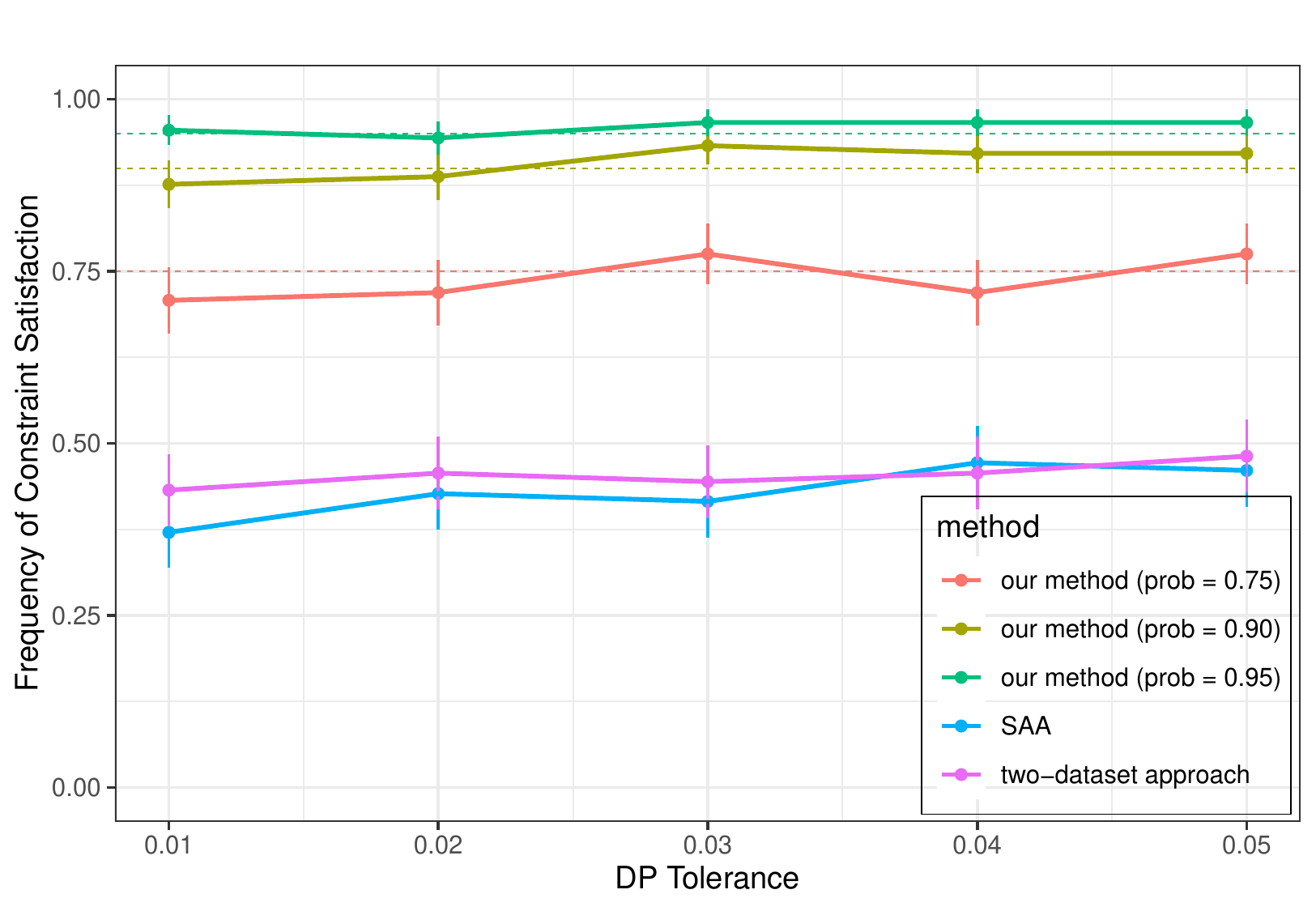}
\end{minipage}
\begin{minipage}{0.48\textwidth}
  \centering
  \includegraphics[width=1\linewidth]{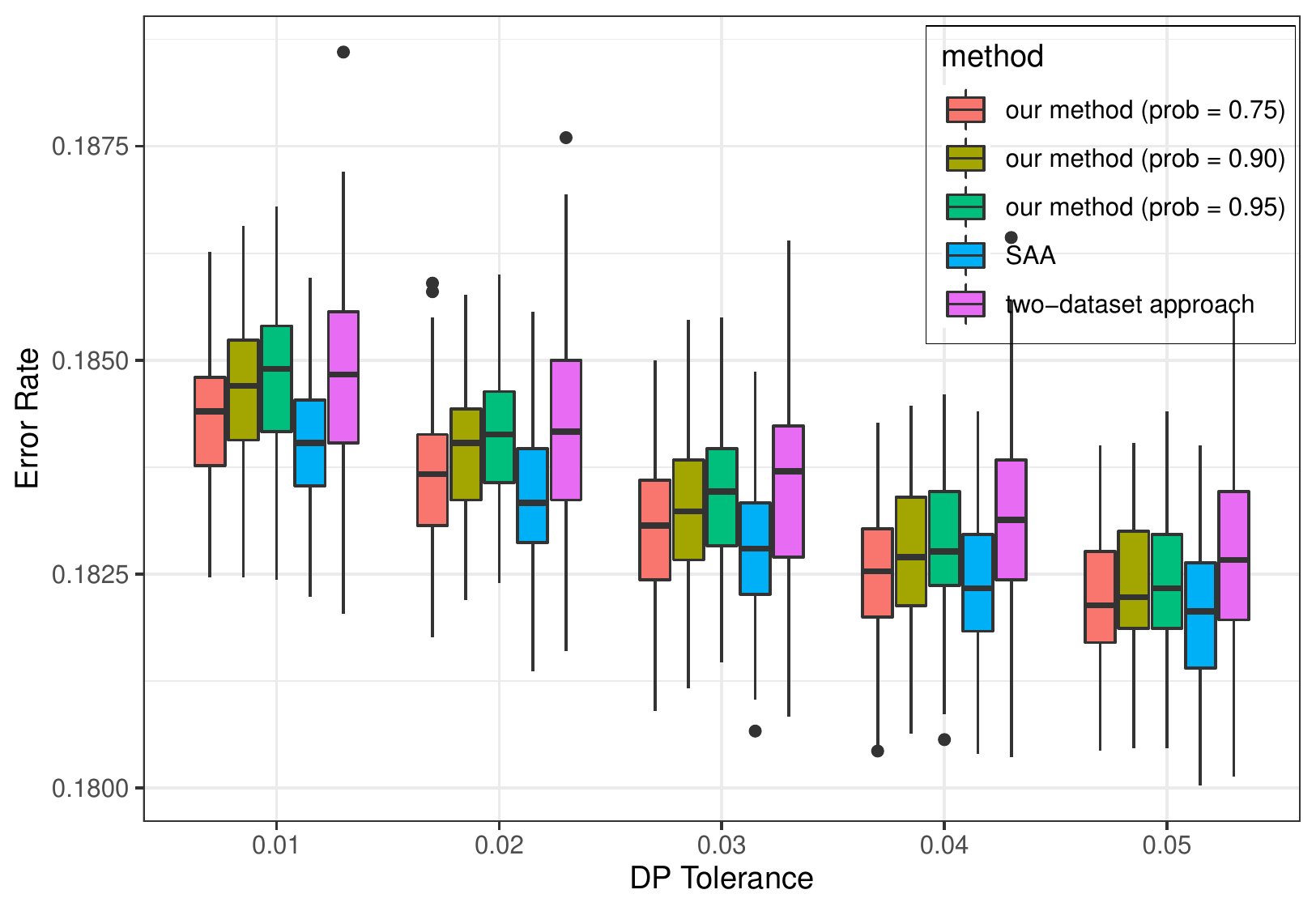}
\end{minipage}
\caption{Frequency of constraint satisfaction (left) and classification error rate (right) for different demographic parity tolerance $\varepsilon \in \{0.01, 0.02, 0.03, 0.04, 0.05\}$. Our methods (with nominal probability $0.75, 0.90,0.95$), sample average approximation (SAA) and two-dataset approach \citep{cotter2019training} are compared.}
\label{fig:8}
\end{figure}

As depicted in Figure \ref{fig:8}, our methods achieve the level of constraint satisfaction frequency specified by the user at the expense of a slight increase in classification error.

\subsection{Credit with $\varepsilon$-true positive rate parity}
Similar to the preceding subsection, now we consider the fairness goal to be $\varepsilon$-true positive rate parity:
\[
\big|\bbP(\hY = 1\mid Y = 1, A = 1) - \bbP(\hY = 1\mid  Y = 1, A = 0)\big|\leq \varepsilon,
\]
where $A = 1$ for individuals with a graduate degree is the advantaged group.

\begin{figure}[h]
\centering
\begin{minipage}{0.48\textwidth}
  \centering
  \includegraphics[width=1\linewidth]{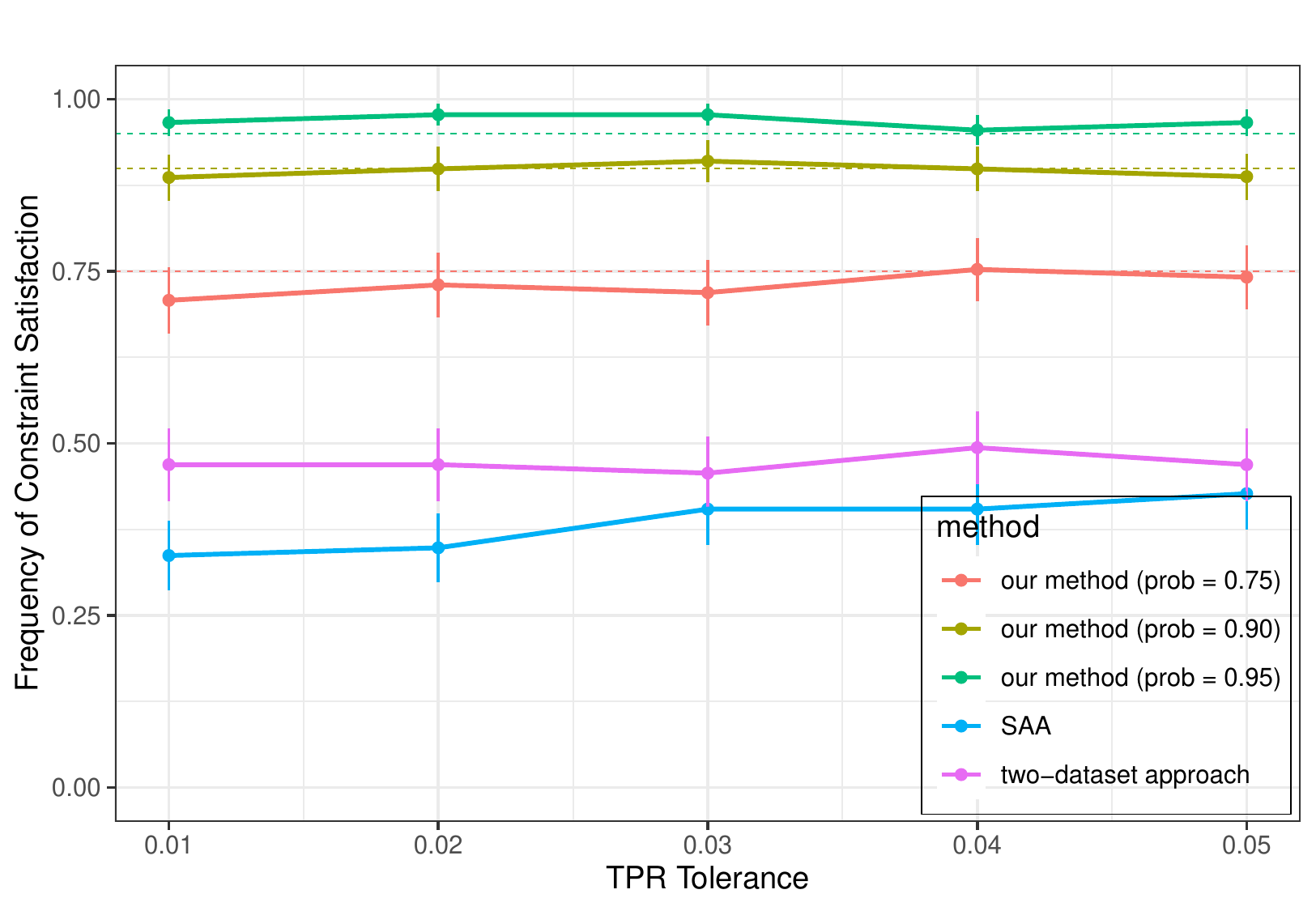}
\end{minipage}
\begin{minipage}{0.48\textwidth}
  \centering
  \includegraphics[width=1\linewidth]{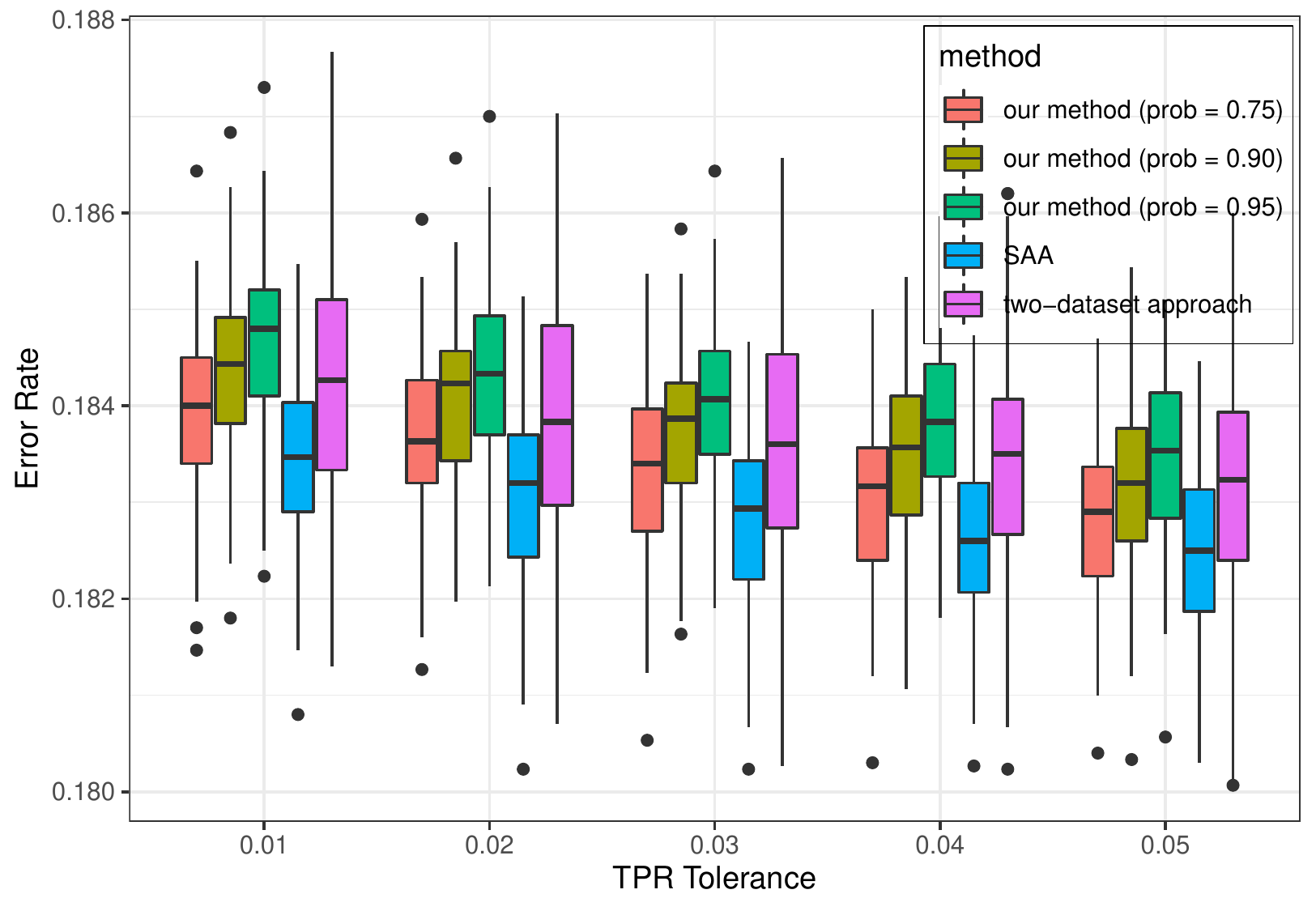}
\end{minipage}
\caption{Frequency of constraint satisfaction (left) and classification error rate (right) for different true positive rate parity tolerance $\varepsilon \in \{0.01, 0.02, 0.03, 0.04, 0.05\}$. Our methods (with nominal probability $0.75, 0.90,0.95$), sample average approximation (SAA) and two-dataset approach \citep{cotter2019training} are compared.}
\label{fig:9}
\end{figure}

The patterns shown in Figure \ref{fig:9} are similar to that in Figure \ref{fig:8}. The SAA has the lowest error rate but the worst generalization of constraint satisfaction. The two-dataset approach increases the SAA's frequency of constraint satisfaction at the cost of an increase in classification error rates. Our methods can achieve the desired high probability of constraint satisfaction, whereas neither of the two baselines can.

